\documentclass[10pt,twocolumn,letterpaper]{article}

\usepackage{cvpr}
\usepackage{times}
\usepackage{epsfig}
\usepackage{graphicx}
\usepackage{amsmath}
\usepackage{amssymb}
\usepackage{mathrsfs}
\usepackage{comment}
\usepackage{colortbl}
\usepackage{color}
\usepackage{subfig}
\usepackage{xcolor}
\usepackage{soul}
\usepackage{pifont}
\usepackage{multirow}
\usepackage{dblfloatfix}

\definecolor{LightBlue}{rgb}{0.8,0.9,1.0}
\definecolor{LightGray}{rgb}{0.9,0.9,0.9}

\usepackage{listings}
\usepackage{color}
\usepackage{capt-of}
\usepackage{multicol}

\definecolor{dkgreen}{rgb}{0,0.6,0}
\definecolor{gray}{rgb}{0.5,0.5,0.5}
\definecolor{mauve}{rgb}{0.58,0,0.82}
\definecolor{codegreen}{rgb}{0.2,0.2,0.2}
\definecolor{codegray}{rgb}{0.5,0.5,0.5}
\definecolor{codepurple}{rgb}{0.58,0,0.82}
\definecolor{backcolour}{rgb}{0.95,0.95,0.95}

\lstdefinestyle{mystyle}{
	backgroundcolor=\color{backcolour},   
	commentstyle=\color{codegreen},
	keywordstyle=\color{magenta},
	numberstyle=\tiny\color{codegray},
	stringstyle=\color{codepurple},
	basicstyle=\footnotesize,
	breakatwhitespace=false,         
	breaklines=true,                 
	captionpos=b,                    
	keepspaces=true,                 
	numbers=left,                    
	numbersep=5pt,                  
	showspaces=false,                
	showstringspaces=false,
	showtabs=false,                  
	tabsize=2
}

\definecolor{ambrish_color}{RGB}{56,108,176} 

\usepackage[pagebackref=true,breaklinks=true,letterpaper=true,colorlinks,bookmarks=false]{hyperref}

\cvprfinalcopy 

\def\cvprPaperID{0830} 

\ifcvprfinal\fi

\makeatletter
\DeclareRobustCommand\onedot{\futurelet\@let@token\@onedot}
\def\@onedot{\ifx\@let@token.\else.\null\fi\xspace}
\def\eg{\emph{e.g}\onedot}

\def\etc{\emph{etc}\onedot} 
 
\def\etal{\emph{et al}\onedot}
\makeatother

\title{Learning to Generate Synthetic Data via Compositing}

\author{
	Shashank Tripathi$^{1,2^{\star\dag}}$  \quad \quad Siddhartha Chandra$^{1^{\star}}$ \quad \quad Amit Agrawal$^1$ \\ \quad \quad Ambrish Tyagi$^1$ \quad \quad James M. Rehg$^1$ \quad \quad Visesh Chari$^1$ \\
	\quad$^1$Amazon Lab126\quad $^2$Carnegie Mellon University\\
	\tt\small \{shatripa, chansidd, aaagrawa, ambrisht, jamerehg, viseshc\}@amazon.com 
}

\begin{document}

\maketitle
\renewcommand*{\thefootnote}{$\star$}
\setcounter{footnote}{1}
\footnotetext{Equal Contribution}
\renewcommand*{\thefootnote}{$\dag$}
\setcounter{footnote}{2}
\footnotetext{Work done during an internship at Amazon Lab126}
\thispagestyle{empty}

\begin{abstract}
		We present TERSE, a task-aware approach to synthetic data generation. 
        Our framework employs a trainable synthesizer network that is optimized to produce meaningful training
        samples by assessing the strengths and weaknesses of a `target' network. The synthesizer and
        target networks are trained in an adversarial manner wherein each network is updated with a
        goal to outdo the other. 
        Additionally, we ensure the synthesizer generates
        realistic data by pairing it with a discriminator trained on real-world images.
        Further, to make the target classifier invariant to blending artefacts, we introduce these artefacts
        to background regions of the training images so the target does not over-fit to them.

        We demonstrate the efficacy of our approach by applying it to different target networks
        including a classification network on AffNIST, and two object detection networks
        (SSD, Faster-RCNN) on different datasets. On the AffNIST benchmark, our approach is able to surpass the
        baseline results with just half the training examples.
        On the VOC person detection benchmark, we show improvements of up to $2.7\%$ 
        as a result of our data augmentation. Similarly on the GMU detection benchmark, we report a performance boost of $3.5\%$
        in mAP over the baseline method, outperforming the previous state of the art
        approaches by up to $7.5\%$ on specific categories.
\end{abstract}

\begin{figure}
  \includegraphics[width=0.5\textwidth,height=2.5in]{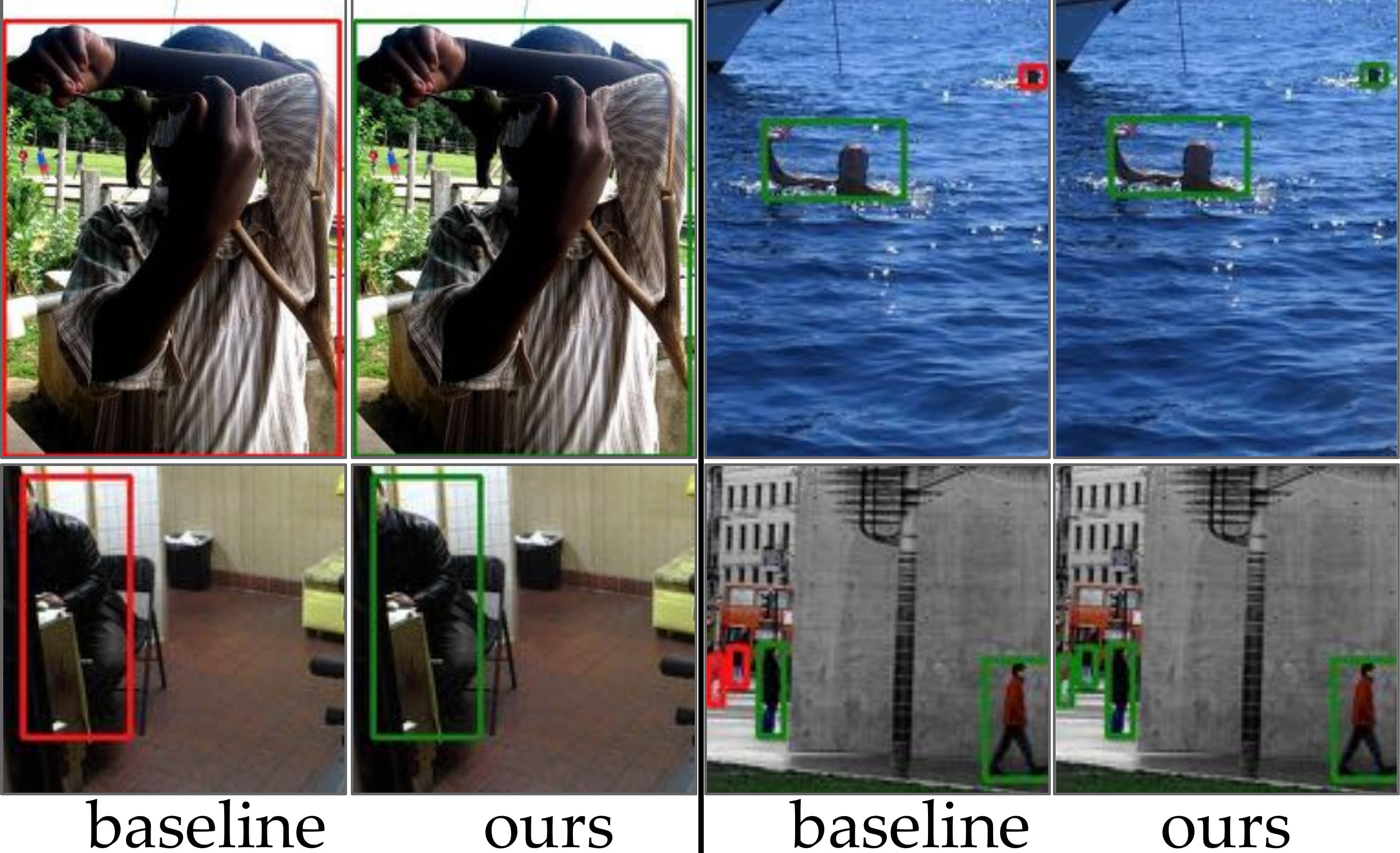}
  \caption{
  Comparison of object detection results using SSD. \emph{Baseline}: trained on VOC data,
    \emph{Ours}: trained on VOC and synthetic data generated using our approach.
      Green and red bounding boxes denote {\color{green}{correct}} and {\color{red}{missed}}
      detections
        respectively. SSD fine-tuned with our synthetic
        data shows improved performance on small, occluded and truncated person instances.}
 %
  \label{fig:teaser-figure}
  \vspace{-0.25in}
\end{figure}

\section{Introduction}
\label{section:introduction}

\begin{figure*}[h!]
	\centering
	\includegraphics[width=\textwidth]{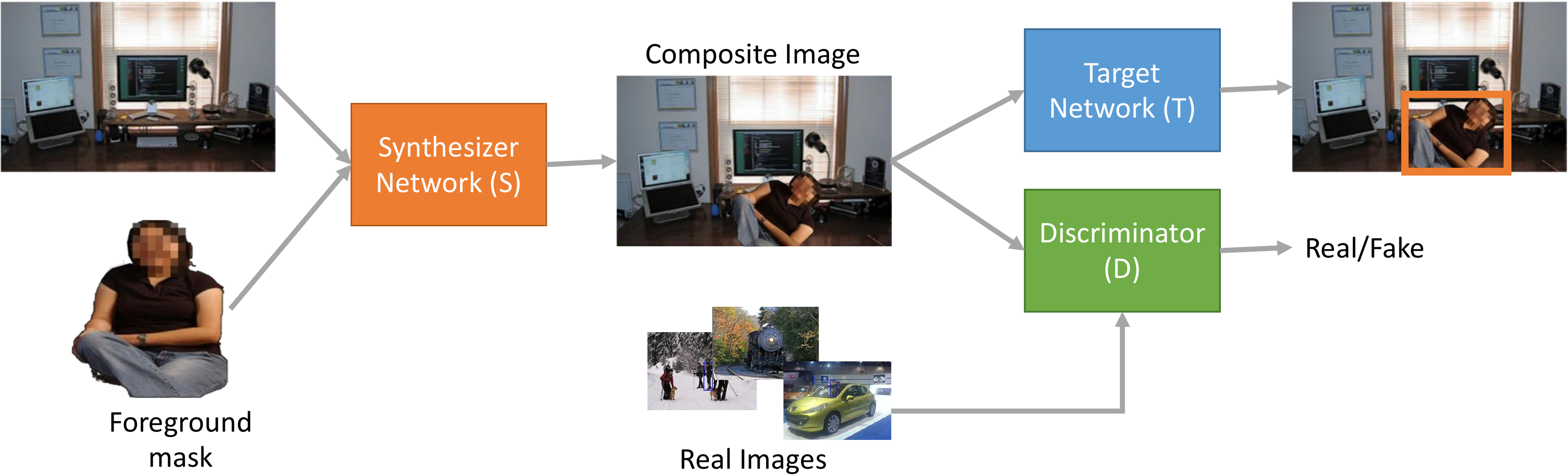}
    \caption{ Our pipeline consists of three components: a synthesizer $\mathcal{S}$, the target
    network $\mathcal{T}$, and
    a natural image discriminator $\mathcal{D}$. $\mathcal{S}$ pastes an optimally transformed
    foreground into a background image to generate a synthetic composite image that can
    `fool' $\mathcal{T}$ and $\mathcal{D}$.
    $\mathcal{T}$ is updated
    using the synthesized image to improve its accuracy. $\mathcal{D}$ provides feedback to
    $\mathcal{S}$ to improve realism of the synthetic image. $\mathcal{T}$ and $\mathcal{D}$ are
    updated with the synthesized data, in lock-step with $\mathcal{S}$. }
    \vspace{-0.2in}
	\label{fig:pipeline1}
\end{figure*}

Synthetic data generation is now increasingly utilized to overcome the burden of creating large
supervised datasets for training deep neural networks. A broad range of data synthesis
approaches have been proposed in literature, ranging from photo-realistic image
rendering~\cite{Krahenbuhl2018,playing,gul} and learning-based image
synthesis~\cite{Salimans2016,apple,Tran2017a} to methods for \emph{data augmentation} that automate the process for generating new example images from an existing training set~\cite{Fawzi2016,Gupta2016a,Hauberg2016,Ratner2017}. Traditional approaches to data augmentation have exploited image transformations that
preserve class labels~\cite{autoaugment,Tran2017a}, while recent works \cite{Hauberg2016,Ratner2017} use a more general set of image transformations, including even compositing images.

For the task of object detection,
recent works have explored a compositing-based approach to data augmentation in which additional
training images are generated by pasting cropped foreground objects on new backgrounds
~\cite{modeling-visual-context,cut-paste-learn,synthesize-detection-indoor}. The compositing approach, which is the
basis for this work, has two main advantages in comparison to image synthesis: 1) the domain gap
between the original and augmented image examples tends to be minimal (resulting primarily from
blending artefacts) and 2) the method is broadly-applicable, as it can be applied to any image
dataset with object annotations.

A limitation of prior approaches is that the process that generates synthetic data is
\emph{decoupled} from the process of training the target classifier. As a consequence, the data
augmentation process may produce many examples which are of little value in improving performance of
the target network.
We posit that a synthetic data generation approach must generate data having three important characteristics. It must
be a) \emph{task-aware}: generate hard examples that help improve target network performance,
b) \emph{efficient}: generate fewer and meaningful data samples,
and c) \emph{realistic}: generate realistic examples that help minimize domain gaps and improve
generalization.

We achieve these goals by developing a novel approach to data synthesis called TERSE, short for Task-aware Efficient Realistic Synthesis of Examples. We set up a 3-way
competition among the synthesizer, target and discriminator networks. The synthesizer is tasked with
generating composite images by combining
a given background with an optimally transformed foreground, such that it can fool the target
network as shown in Figure~\ref{fig:pipeline1}. The goal of the target network is
to correctly classify/detect all instances of foreground object in the composite images.
The synthesizer and target networks are updated iteratively, in a lock-step.
We additionally introduce a real image discriminator to ensure the composite images generated by the
synthesizer conform to the real image distribution. Enforcing realism prevents the model from
generating artificial examples which are unlikely to occur in real images, thereby improving the
generalization of the target network.

A key challenge with all composition-based methods is the sensitivity of trained models to blending artefacts.  
The target and discriminator networks can easily learn to latch on to the blending artefacts, thereby rendering the data generation process ineffective.
To address these issues with blending, Dwibedi~\etal~\cite{cut-paste-learn}
employed $5$ different blending methods so that the target network does not over-fit to a particular
blending artefact. We propose an alternate solution to this problem by synthesizing
examples that contain similar blending artefacts in the background. The artefacts are generated by
pasting foreground shaped cutouts in the background images. This makes the target network insensitive to any
blending artefacts around foreground objects, since the same artefacts are present in the background images as well.

We apply our synthesis pipeline to demonstrate improvements on tasks including digit classification on
the AffNIST dataset~\cite{affnist}, object localization using SSD~\cite{SSD-network} on Pascal
VOC~\cite{Pascal-voc}, and instance detection using Faster RCNN~\cite{FasterRCNN} on GMU
Kitchen~\cite{GMU-kitchen-dataset}  dataset. 
We demonstrate that our approach is a) \emph{efficient}: we achieve similar performance to
baseline classifiers 
using less than $50\%$ data~(Sec.~\ref{subsec:affnist}), b) \emph{task-aware}:
networks trained on our data achieve up to $2.7\%$ improvement for person detection~(Sec.~\ref{subsection:pascal_VOC}) and $3.5\%$
increase in mAP over all classes on the GMU kitchen dataset over
baseline~(Sec.~\ref{subsection:gmu_kitchen}). We also show that our approach produces $>$2X hard positives compared to
state-of-the-art~\cite{modeling-visual-context,cut-paste-learn} for person detection.
Our paper makes the following contributions:
\begin{itemize}
    \setlength{\itemsep}{0pt}
    \setlength{\leftmargin}{0pt}
    \item We present a novel image synthesizer network that learns to create composites 
    specifically to \emph{fool} a target network. We show that the synthesizer is effective at
    producing hard examples to improve the target network.
	
    \item We propose a strategy to make the target network invariant to artefacts in the
      synthesized images, by generating additional hallucinated artefacts in the background images.
	
    \item We demonstrate applicability of our framework to image classification, object detection,
      and instance detection.
\end{itemize}

\begin{figure}[h!]
  \includegraphics[width=0.5\textwidth]{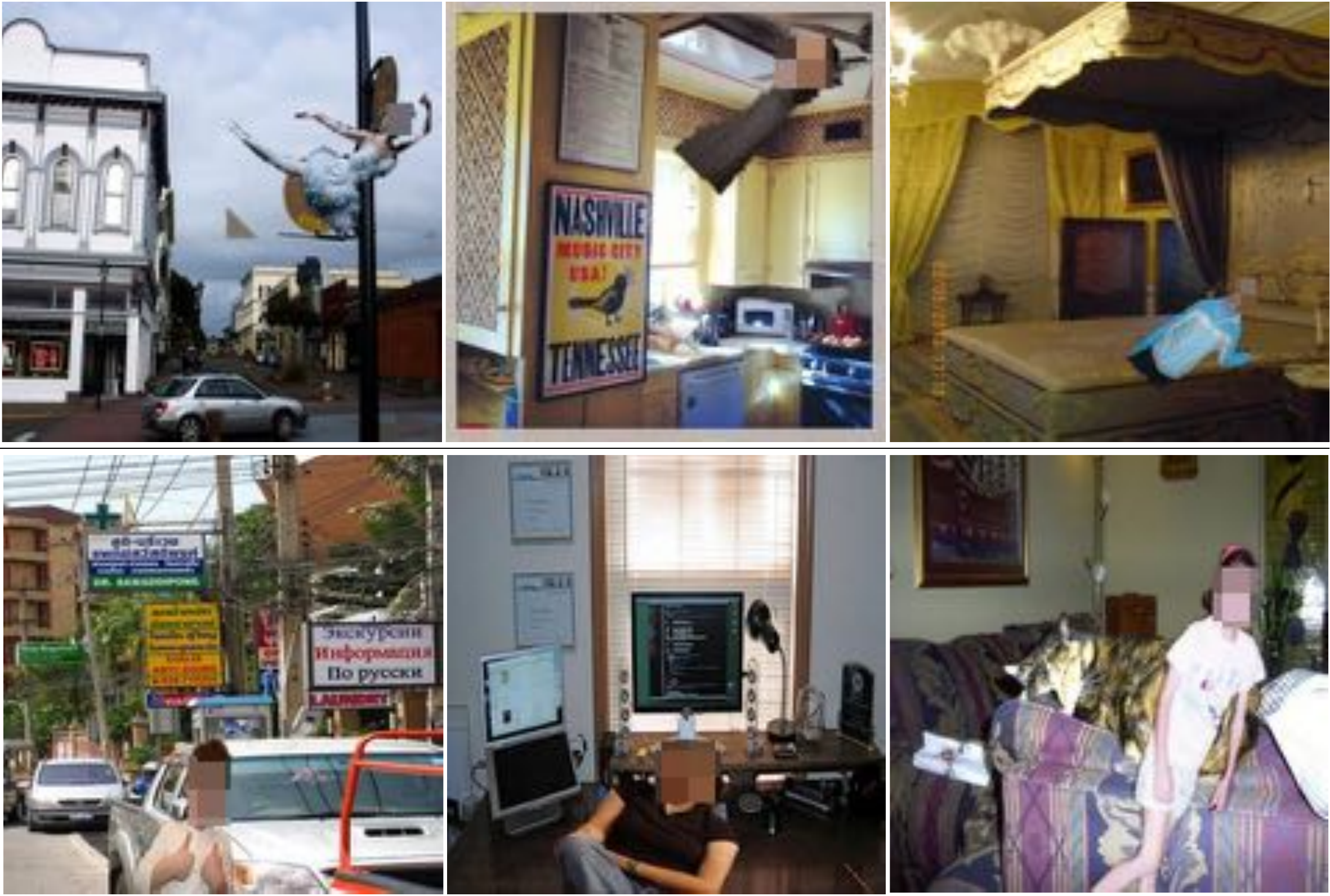}
	\caption{Using a discriminator improves the realism of generated images. (Top) Composite images
    generated without discriminator in the loop. (Bottom) Composite images generate with
    discriminator.}
	\label{fig:sleep_data_without_discriminator}
    \vspace{-0.25in}
\end{figure}

\section{Related Work}
\label{section:related_work}

To the best of our knowledge, ours is the first approach to generate synthetic data by compositing
images in a task-aware fashion.
Prior work on synthetic data generation can be organized into three groups: 1) Image composition,
2) Adversarial generation, and 3) Rendering.

\textbf{Image Composition:} Our work is inspired by recent cut and paste
approaches~\cite{modeling-visual-context,cut-paste-learn,synthesize-detection-indoor} to synthesize
positive examples for object detection tasks. The advantage of these approaches comes from
generating novel and diverse juxtapositions of foregrounds and backgrounds that can substantially 
increase the available training data.
The starting point for our work is the approach of Dwibedi~\etal~\cite{cut-paste-learn}, who were
first to demonstrate empirical boosts in performance through the cut and paste procedure.
Their approach uses random sampling to decide the placement of foreground patches on background
images. However, it can produce unrealistic compositions which
limits generalization performance as shown by~\cite{modeling-visual-context}. To help with generalization, prior
works~\cite{modeling-visual-context,synthesize-detection-indoor} exploited contextual
cues~\cite{Divvala2009,Mottaghi2014,Oliva2007} to guide the placement of foreground patches and
improve the realism of the generated examples. Our data generator network implicitly encodes
contextual cues which is used to generate realistic positive examples, guided by the discriminator. We therefore avoid the need to construct explicit models of
context~\cite{Divvala2009,modeling-visual-context}. 
Other works have used image compositing to improve image synthesis~\cite{when-where-who},
multi-target tracking~\cite{lucid}, and pose tracking~\cite{synthetic-occlusion-augmentation}.
However, unlike our approach, none of these prior works optimize for the target network while generating synthetic data.
%

\textbf{Adversarial learning:} Adversarial learning has emerged as a powerful framework for tasks such as image synthesis, generative sampling, synthetic data generation~\etc~\cite{chen20193d, drover2018can, stgan, song2018constructing} We employ an adversarial learning paradigm to train our
synthesizer, target, and discriminator networks. Previous works such as A-Fast-RCNN
~\cite{Wang2017a} and the adversarial spatial transformer (ST-GAN)~\cite{stgan} have also employed
adversarial learning for data generation. 
The A-Fast-RCNN method uses adversarial spatial dropout to simulate
occlusions and an adversarial spatial transformer network to simulate object deformations, but does
not generate new training samples.
The ST-GAN approach uses a generative model to synthesize realistic composite images, but does not
optimize for a target network. 

\textbf{Rendering:} Recent works~\cite{sintel,pre-trained-features-synthetic,playing,apple,domain-randomization,facebook-house} have used simulation engines to
render synthetic images to augment training data. Such approaches
allow fine-grained control over the scale, pose, and spatial positions of foreground objects,
thereby alleviating the need for manual annotations.
A key problem of rendering based approaches is the domain difference between synthetic and real
data. Typically, domain adaptation algorithms~(\eg~\cite{apple}) are necessary to bridge this gap.
However, we avoid this problem by compositing images only using real data. 

\textbf{Hard example mining:}
Previous works have shown the importance of hard examples for training robust models
\cite{unsupervised-hard-negative-mining,focal-loss,shrivastava2016training,spatially-transformed-adversarial,lrm,SSD-network}.
However, most of these approaches mine existing training data to identify hard examples and are
bound by limitations of the training set. Unlike our approach, these methods do not generate new examples.
Recently, \cite{ICN,zhao2018towards} proposed data augmentation for generating transformations that yields
additional pseudo-negative training examples. In contrast, we generate hard positive examples.
 

%
%
%
%

\section{TERSE Data Synthesis}
\label{section:rem_model}



Our approach for generating hard training examples through image composition requires as input a background image, $b$, and a
segmented foreground object mask, $m$, from the object classes of interest. The learning problem is
formulated as a 3-way competition among the synthesizer $\mathcal{S}$, the target $\mathcal{T}$, and
the discriminator $\mathcal{D}$. We optimize $\mathcal{S}$ to produce composite images that can fool
both $\mathcal{T}$ and $\mathcal{D}$. $\mathcal{T}$ is updated with the goal to
optimize its target loss function, while $\mathcal{D}$ continues to improve its classification
accuracy. The resulting synthetic images are both realistic and constitute hard examples for $\mathcal{T}$.
The following sections describe our data synthesis pipeline and end-to-end training process in more detail.
	
\subsection{Synthesizer Network} 
The synthesizer operates on the inputs $b$ and $m$ and outputs a transformation function,
$A$. This transformation is applied to the foreground mask to produce a composite synthetic image, $f = b \oplus
A(m)$, where $\oplus$ denotes the alpha-blending~\cite{stgan} operation. In this work, we restrict
$A$ to the set of 2D affine transformations (parameterized by a $6-$ dimensional feature vector), but the approach can trivially be extended to other
classes of image transformations. $b,f,A$  are then fed to a Spatial Transformer module \cite{spatial-transformer-networks} which produces the composite image $f$ (Figure ~\ref{fig:pipeline1}). The composite image is fed to the discriminator and target networks with the goal of fooling both of them.
The synthesizer is trained in lockstep with the target and discriminator as described in the following sections.



\begin{figure}
  \centering
  \includegraphics[width=3in,height=2in]{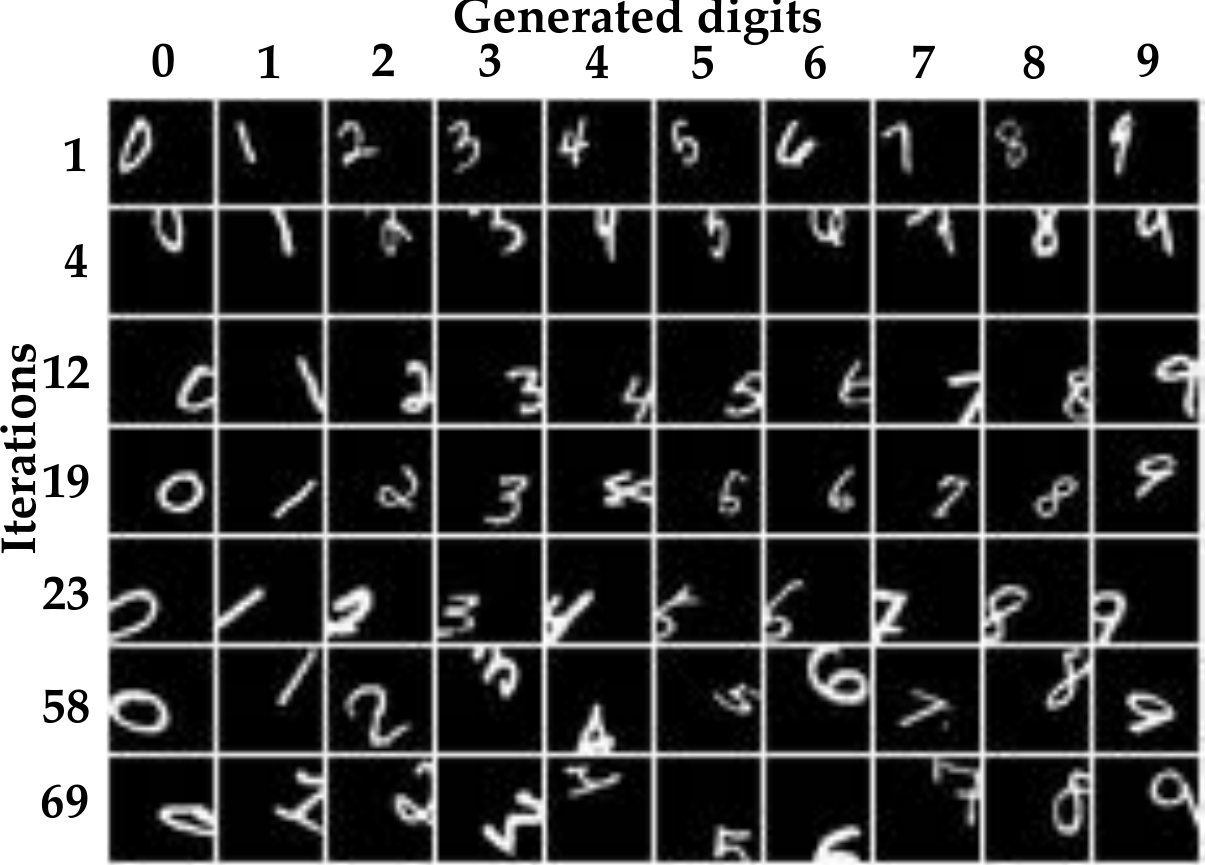}
  \caption{Data generated by our approach over multiple iterations for the AffNIST experiment
    (Section~\ref{subsec:affnist}). As training progresses (top to bottom) the synthesized examples
    become more complex, from single modes of failure of the target network to multiple modes at
    later stages.}
  \vspace{-0.25in}
  \label{fig:affnist-progression}
\end{figure}

\noindent\textbf{Blending Artefacts:} In order to paste foreground regions into backgrounds, we use the standard alpha-blending method described in
\cite{spatial-transformer-networks}. One practical challenge, as discussed in \cite{cut-paste-learn}, is that the target model can
learn to exploit any artefacts introduced by the blending function, as these will always be associated with positive examples, thereby harming the generalization of the classifier. Multiple blending strategies are used in~\cite{cut-paste-learn} to discourage the
target model from exploiting the blending artefacts. However, a target model with sufficient
capacity could still manage to over-fit on all of the different blending functions that were used. Moreover, it is challenging to generate a large number of candidate blending functions due to the need to ensure differentiability in end-to-end learning.

\begin{figure}
  \centering
  \includegraphics[width=0.5\textwidth]{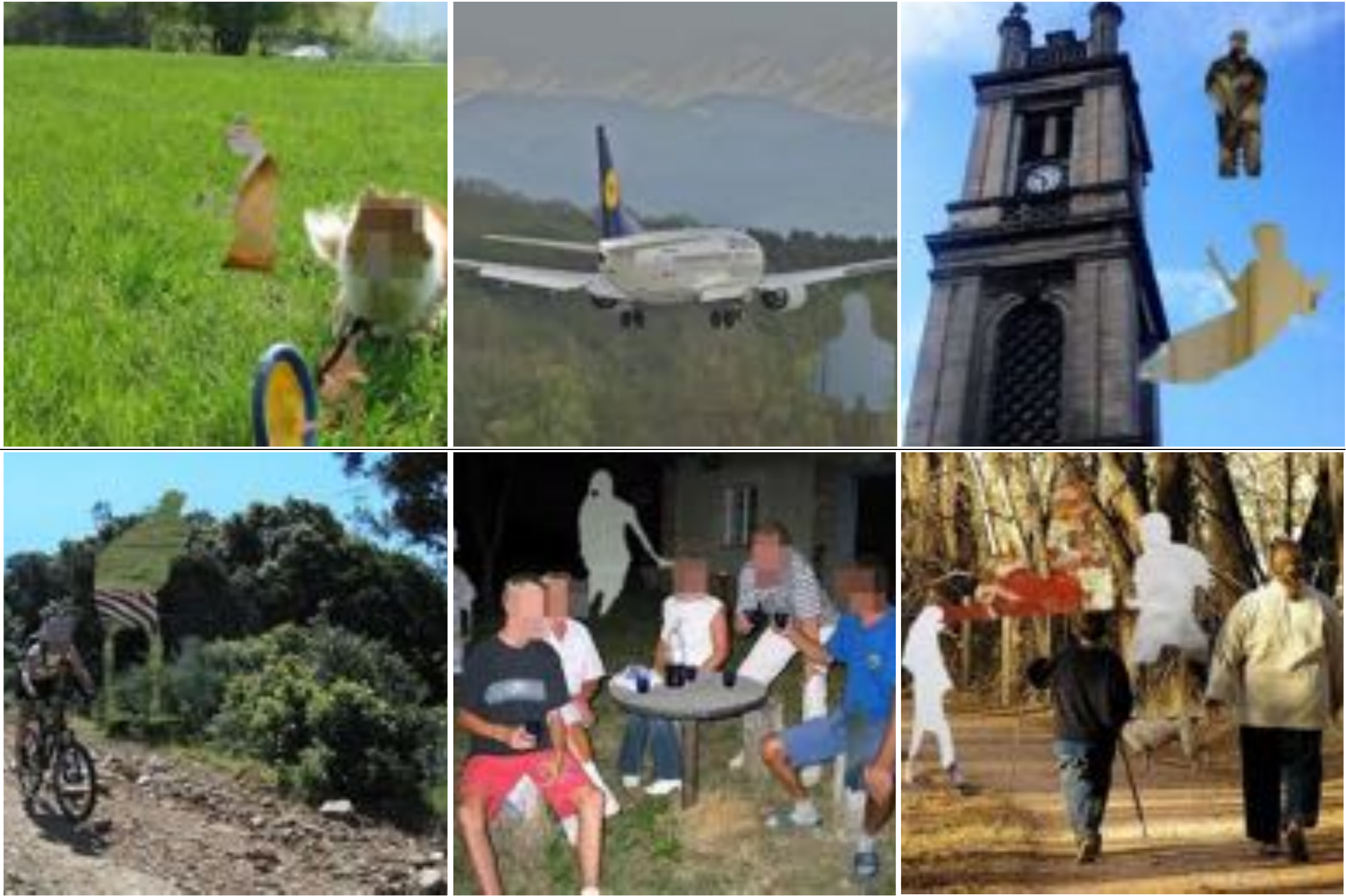}
  \caption{ Examples of blending artefacts pasted into \emph{background regions} of training
  images in order to remove any discriminative cues associated with compositing.
  A random foreground-shaped cut-out from a different background image is pasted on the background
  regions of the given image. Images from COCO (top row) and VOC (bottom row) are shown.}
  \label{fig:blending_artefacts}
\end{figure}

We propose a simple and effective strategy to address this problem. We
explicitly introduce blending artefacts into the background regions of synthesized images (see
Fig.~\ref{fig:blending_artefacts}). To implement this strategy, we (i) randomly choose
a foreground mask from our training set, (ii) copy background region shaped like this mask from one
image, and (iii) paste it onto the background region in another image using the same blending
function used by $\mathcal{S}$.
As a consequence of this process, the presence of a composited region in an image no longer has any discriminative value, 
as the region could consist of either foreground or background pixels.
This simple strategy makes both the discriminator and the target model invariant to
any blending artefacts. 

\subsection{Target Network} 

The target model is a neural network trained for specific objectives such as
image classification, object detection, semantic segmentation, regression,~\etc. Typically, we first
train the target $\mathcal{T}$ with a labeled dataset to obtain a baseline
level of performance. This pre-trained baseline model $\mathcal{T}$ is then fine-tuned in lockstep with $\mathcal{S}$ and
$\mathcal{D}$.
Our synthetic data generation framework is applicable to a wide range of target networks. Here we derive
the loss for the two common cases of image classification and object detection.

\noindent\textbf{Image Classification:} For the task of image classification, the target loss function $\mathcal{L}_{\mathcal{T}}$ is the standard cross-entropy loss over the training dataset. 


\noindent\textbf{Object Detection:} For detection frameworks such as SSD~\cite{SSD-network} and faster-RCNN~\cite{FasterRCNN}, for each bounding-box proposal, the target
network outputs (a) probability distribution $p = (p^0,\cdots,p^L)$ over the $L+1$ classes in the dataset (including background), (b) bounding-box regression offsets $r \in \mathbb{R}^4$. While SSD uses fixed anchor-boxes, faster-RCNN uses CNN based proposals for bounding boxes. The ground
truth class labels and bounding box offsets for each proposal are denoted by $c$ and $v$,
respectively.
Anchor boxes with an Intersection-over-Union (IoU) overlap greater than $0.5$ with
the ground-truth bounding box are labeled with the class of the bounding box, and the rest are
assigned to the background class. The object detector target $\mathcal{T}$ is trained to optimize the following loss function:
\begin{equation}
\centering
\mathcal{L}_{\mathcal{T}} (p,c,r,v) = \underbrace{-\log(p^{c})}_{\text{classification objective}} + \underbrace{\lambda[c>0] L_{loc}(r,v)}_{\text{localization objective}}
\label{eqn:det}
\end{equation}
where, $L_{loc}$ is the smooth $L_1$ loss function defined in \cite{fast-rcnn}. The Iverson bracket
indicator function $[c>0]$ evaluates to $1$ for $c>0$, i.e. for non-background classes and $0$
otherwise. In other words, only the non-background anchor boxes contribute to the localization
objective.


\subsection{Natural Image Discriminator}
\label{subsection:natural_image_discriminator}
An unconstrained cut-paste approach to data augmentation can produce non-realistic composite
images (see for example Fig.~\ref{fig:sleep_data_without_discriminator}). Synthetic data generated in such a way can still potentially improve the target network as
shown by Dwibedi~\etal~\cite{cut-paste-learn}. However, as others~\cite{Divvala2009,Mottaghi2014,Oliva2007} have shown, generating contextually salient and
realistic synthetic data can help the target network to learn more efficiently and generalize more effectively to real world tasks. 


Instead of learning specific context and affordance models, as employed in aforementioned works, we adopt an adversarial training approach and feed the output of the synthesizer to a discriminator network as negative examples. The discriminator also receives positive examples in the form of real-world images. It acts as a binary
classifier that differentiates between real images $r$ and composite images $f$. For an image $I$,
the discriminator outputs $\mathcal{D}(I)$, i.e. the probability of $I$ being a real image.
$\mathcal{D}$ is trained to maximize the following objective:
\begin{equation}
\mathcal{L}_{\mathcal{D}}  = \mathbb{E}_{r} \log(\mathcal{D}(r)) + \mathbb{E}_f \log(1 - \mathcal{D}(f)).
\label{eqn:discriminator}
\end{equation}

As illustrated in Figure~\ref{fig:sleep_data_without_discriminator}, the discriminator helps
the synthesizer to produce more natural looking images.

\subsection{Training Details}
The three networks, $\mathcal{S}$, $\mathcal{T}$, and $\mathcal{D}$, are trained according to the following objective function:
\begin{equation}
\mathcal{L}_{\mathcal{S}, \mathcal{T}, \mathcal{D}} =  \max_{\mathcal{S}} \min_{\mathcal{T}} \mathcal{L}_{\mathcal{T}} +
\min_{\mathcal{S}} \max_{\mathcal{D}} \mathcal{L}_{\mathcal{D}}
\label{eqn:final_equation_classification}
\end{equation}
For a given training batch, parameters of $\mathcal{S}$ are updated while keeping parameters of $\mathcal{T}$ and $\mathcal{D}$
fixed. Similarly, parameters of $\mathcal{T}$ and $\mathcal{D}$ are updated by keeping parameters of 
$\mathcal{S}$ fixed. $\mathcal{S}$ can be seen as an adversary to both $\mathcal{T}$ and $\mathcal{D}$.



\begin{figure}
  \centering
  \includegraphics[width=0.45\textwidth]{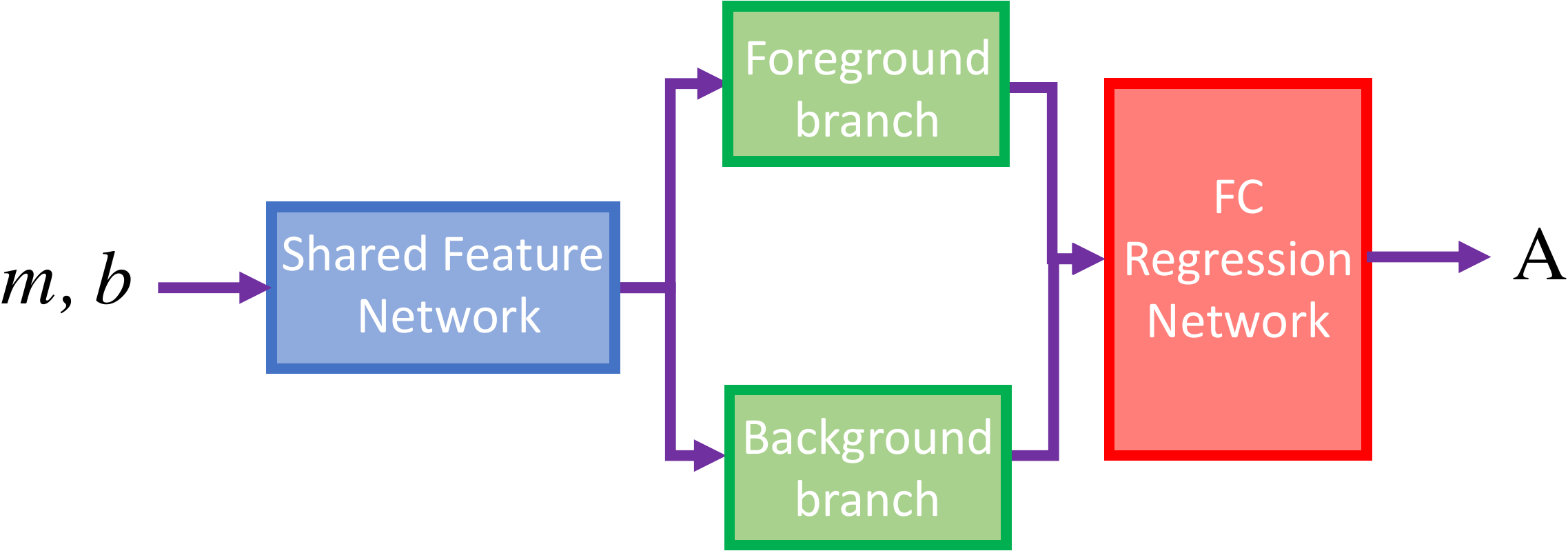}
  \caption{Synthesizer Architecture}
  \label{fig:synthesizer_architecture}
\end{figure}
\noindent \textbf{Synthesizer Architecture.} Our synthesizer network
(Figure~\ref{fig:synthesizer_architecture}) consists of (i) a shared low-level feature extraction backbone that performs identical
feature extraction on foreground masks $m$ and background images $b$, (ii) and parallel branches for mid-level feature extraction on $m,b$, and (iii)
a fully-connected regression network that takes as input the concatenation of mid-level features of $m,b$ and outputs a $6-$dimensional feature vector representing
the affine transformation parameters. For the AffNIST experiments, we use a $2-$ layer network as the
backbone. For experiments on Pascal VOC and GMU datasets, we use the VGG-16~\cite{vgg} network up to
\emph{Conv-5}. The mid-level feature branches each consist of $2$ bottlenecks, with one convolutional layer, followed by ReLU and BatchNorm layers. The regression network consists of $2$ convolutional and $2$ fully connected layers.\\

\noindent \textbf{Synthesizer hyper parameters.} We use Adam~\cite{adam} optimizer with a learning rate of
$1e-3$ for experiments on the AffNIST dataset and $1e-4$ for all other experiments. We set the
weight decay to $0.0005$ in all of our results.

\noindent \textbf{Target fine-tuning hyper parameters.} For the AffNIST benchmark, the target classifier is finetuned using the SGD optimizer with a learning rate of $1e-2$, a momentum of $0.9$ and weight decay of $0.0005$. For person detection on VOC, the SSD is finetuned using the Adam optimizer with a learning rate of $1e-5$, and weight decay of $0.0005$. For experiments on the GMU dataset, the faster-RCNN model is finetuned using the SGD optimizer with a learning rate of $1e-3$, weight decay of $0.0001$ and momentum of $0.9$.

\section{Experiments \& Results}
\label{section:evaluation}

\begin{figure}
  \centering
  \includegraphics[width=0.45\textwidth,height=45mm]{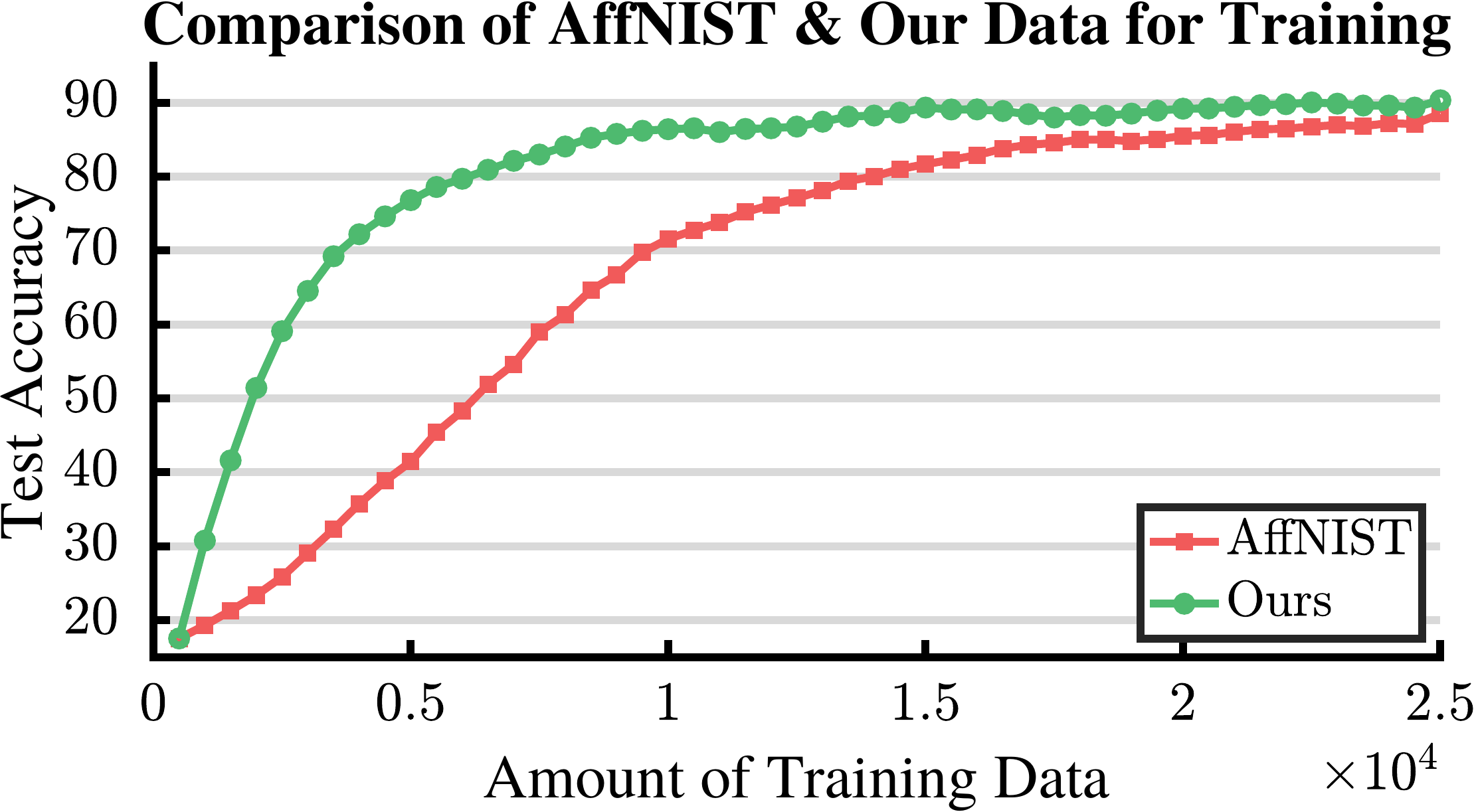} \\
  \captionof{figure}{Performance of MNIST classifier on AffNIST test data when progressively augmented
  with (i) AffNIST training data (red), (ii) our synthetic images (green). 
  Our approaches achieves baseline accuracy ($\approx 90\%$) with less than half the data ($12K$ samples vs
  $25K$ samples). Note that even with $5K$ samples we reach an accuracy of $\approx 80\%$, compared to
  baseline accuracy of $\approx 40\%$.
}
\label{fig:affnist-result}
\captionof{table}{Our approach achieves better classification accuracy compared to previous pseudo-negative data synthesis approaches
on AffNIST dataset. Numbers are reported from the respective papers.}
 \begin{tabular}{|p{1.5cm}|p{0.65cm}|p{0.65cm}|p{0.65cm}|p{0.65cm}|p{0.65cm}|p{0.55cm}|}
    \hline
    Method & \rotatebox{90}{DCGAN\cite{DCGAN}} &
    \rotatebox{90}{WGAN-GP\cite{WCGAN-GP}} &
    \rotatebox{90}{ICN\cite{ICN}} & \rotatebox{90}{WINN\cite{WINN}} &
    \rotatebox{90}{ITN\cite{zhao2018towards}} & Ours \\
      \hline
      Error (\%) & 2.78 & 2.76 & 2.97 & 2.56 & 1.52 & \textbf{0.99} \\
      \hline
  \end{tabular}
  \label{tab:affnist-comparison}
\end{figure}

We now present qualitative and quantitative results to demonstrate the efficacy of our data synthesis approach.

\subsection{Experiments on AffNIST Data}
\label{subsec:affnist}
We show the \emph{efficiency} of data generated using our approach on
AffNIST~\cite{affnist} hand-written character dataset.
It is generated by transforming
MNIST~\cite{mnist} digits by randomly sampled affine transformations. 
For generating synthetic images with our framework, 
we apply affine transformations on MNIST digits and paste them onto black background images.

\noindent \textbf{Target Architecture:}
The target classification model is a neural network consisting of 
two $5 \times 5$ convolutional layers with $10$ and $20$ output channels, respectively. Each layer 
uses ReLU activation, followed by a dropout layer. The output features are then processed by two fully-connected layers with 
output sizes of $50$ and $10$, respectively. 


We conducted two experiments with AffNIST dataset:
\noindent \textbf{Efficient Data Generation:} The baseline classifier is trained on MNIST. The AffNIST
model is fine-tuned by incrementally adding samples undergoing random affine transformation as
described in~\cite{affnist}. Similarly, results from \emph{our} method incrementally improves the
classifier using composite images generated by $\mathcal{S}$.

Figure~\ref{fig:affnist-result} shows the performance of the target model on the AffNIST test set by progressively increasing the
size of training set. When trained on MNIST dataset alone, the target model has a classification accuracy of $~17\%$ on
the AffNIST test set. We iteratively fine-tune the MNIST model from this point by augmenting
the training set with $500$ images either from the AffNIST training set (red curve) 
or from the synthetic images generated by $\mathcal{S}$ (green curve). 
Note that our approach achieves baseline accuracy with less than half the data. In addition, as
shown in Figure~\ref{fig:affnist-result}, using only $5K$ examples, our method improves accuracy
from $40\%$ to $80\%$. Qualitative results in Figure~\ref{fig:affnist-progression} shows the progression of examples
generated by $\mathcal{S}$. As training progresses, our approach generates increasingly hard
examples in a variety of modes. 

\noindent \textbf{Improvement in Accuracy:}
In Table~\ref{tab:affnist-comparison}, we compare our approach with recent methods~\cite{zhao2018towards,DCGAN,WCGAN-GP,WINN,ICN} that generate synthetic data to
improve accuracy on AffNIST data. For the result
in Table~\ref{tab:affnist-comparison}, we use $55000$, $5000$, $10000$ split for training,
validation and testing as in~\cite{zhao2018towards} along with the same classifier architecture. We outperform
hard negative generation approaches~\cite{zhao2018towards,WINN,ICN} by achieving a low error rate
of $0.99\%$. Please find more details in the supplementary.


\subsection{Experiments on Pascal VOC Data}
\label{subsection:pascal_VOC}
We demonstrate improved results using our approach for person detection on
the Pascal VOC dataset~\cite{Pascal-voc}, using the SSD$-300$ network~\cite{SSD-network}. We use
ground-truth person segmentations and bounding box annotations to recover instance masks from VOC 2007 and 2012 training and validation sets as foreground.
Background images were obtained from the COCO dataset~\cite{COCO}. We do an initial clean up of those annotations since we find that 
for about $10\%$ of the images, the annotated segmentations and bounding-boxes do not agree. 
For evaluation we augment the VOC 2007 and
2012 training dataset with our synthetic images, and report mAP for detection on VOC 2007 test set
for all experiments. 
%
%

 \begin{figure}[t!]
  \includegraphics[width=0.5\textwidth]{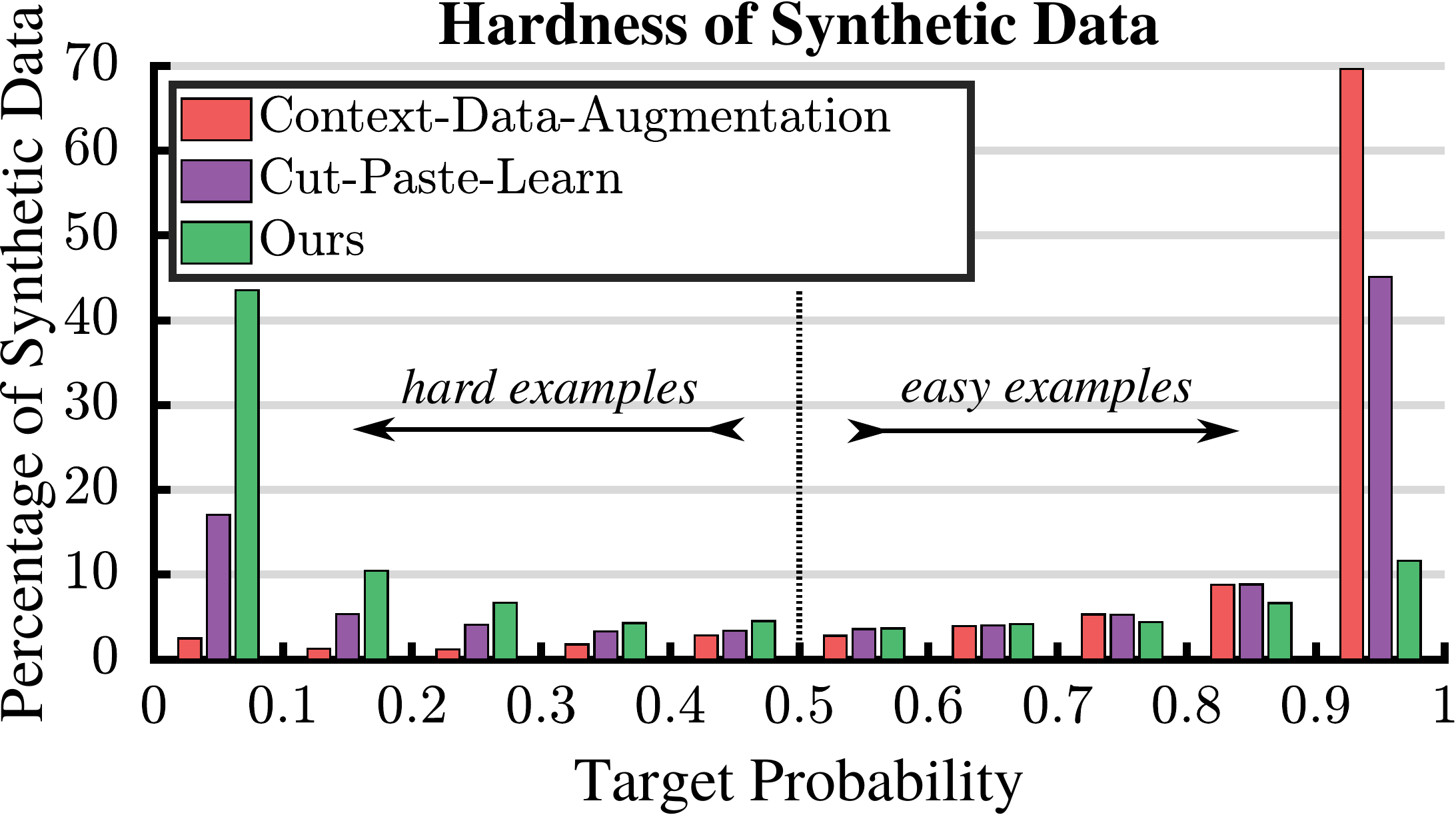}
  \caption{Comparison of our approach with Cut-Paste-Learn~\cite{cut-paste-learn} and
  Context-Data-Augmentation~\cite{modeling-visual-context}, on
  the fraction of hard positives generated for the \emph{person} class.}
  \label{fig:hardness_of_data}
\end{figure}

\newcommand{\cmark}{\ding{51}}%
\newcommand{\xmark}{\ding{55}}%

\begin{table}
\centering
\scalebox{0.85}{
\begin{tabular}{c|p{0.65cm}|p{0.65cm}|p{0.65cm}|p{0.65cm}|p{0.65cm}|p{0.65cm}}\hline
  \multicolumn{3}{c|}{Baseline~\cite{SSD-network}} & \multicolumn{2}{c|}{AP$_{0.5}$ $\rightarrow$ $78.93$} & \multicolumn{2}{c}{AP$_{0.8}$ $\rightarrow$  $29.52$} \\ \hline
Column No. & 1 & 2 & 3 & 4 & 5 & 6 \\ \hline
Ann. Cleanup &   & \cmark & \cmark & \cmark & \cmark & \cmark \\ \hline
Dropout      &   &   & \cmark & \cmark & \cmark & \cmark \\ \hline
Blending     &   &   &   & \cmark & \cmark & \cmark \\ \hline
$1:1$ Ratio  &   &   &   &   & \cmark & \cmark \\ \hline
Discriminator&   &   &   &   &   & \cmark \\ \hline
AP$_{0.5}$   & $79.02$ & $79.13$ & $79.02$ & $79.34$ & $79.61$ & $79.53$ \\ \hline
AP$_{0.8}$   & $29.64$ & $30.72$ & $30.80$ & $31.25$ & $31.96$ & $32.22$ \\ \hline
\end{tabular}}
\caption{Ablation Studies. We show the effect of design choices on the performance of our approach.
Significant improvements are observed by introducing blending
artefacts in background regions (col. 4) and maintaining a $1:1$ ratio between real and synthetic
images (col 5) during training. Adding a discriminator provides additional boost at
AP$_{0.8}$.}
\label{table:ablation_studies}
\end{table}

 \begin{table*}[t!]
  \begin{small}
    \begin{tabular}{p{3.5cm}p{0.7cm}p{0.7cm}p{0.7cm}p{0.7cm}p{0.9cm}p{0.7cm}p{0.7cm}p{0.9cm}p{0.7cm}p{0.7cm}p{0.7cm}p{0.5cm}}
      \hline
    Dataset & coca cola & coffee mate & honey bunches & hunt's sauce & mahatma rice & nature v1 &
    nature v2 & palmolive orange & pop secret & pringles bbq & red bull & mAP \\
    \hline
    Baseline faster-RCNN & $81.9$ & $95.3$ & $92.0$ & $87.3$ & $86.5$ & $96.8$ & $88.9$ & $80.5$ & $92.3$ & $88.9$ & $58.6$ & $86.3$ \\
    \hline
    Cut-Paste-Learn \cite{cut-paste-learn} & $\mathbf{88.5}$ & $95.5$ & $\mathbf{94.1}$ & $88.1$ &
    $\mathbf{90.3}$ & $\mathbf{97.2}$ & $91.8$ & $80.1$ & $94.0$ & $\mathbf{92.2}$ & $65.4$ & $88.8$ \\
    \hline
    Ours & $86.9$ & $\mathbf{95.9}$ & $93.9$ & $\mathbf{90.2}$ & $90.0$ & $96.6$ &
    $\mathbf{92.0}$ & $\mathbf{87.6}$ & $\mathbf{94.9}$ & $90.9$ & $\mathbf{69.2}$ & $\mathbf{89.8}$ \\ \hline
  \end{tabular}
\end{small}
\caption{Comparison of our approach with the baseline Faster-RCNN and~\cite{cut-paste-learn} on the GMU Kitchen Dataset. 
Our approach improves overall mAP and outperforms other approaches in most classes.}
\label{table:gmu_results}
\end{table*}

\subsubsection{Comparison with Previous Cut-Paste Methods}
We compare our results with the performance of the baseline SSD network after fine-tuning it with the data generated by recent approaches
from \cite{modeling-visual-context,cut-paste-learn}. We use the publicly available
software from authors of \cite{modeling-visual-context,cut-paste-learn} to generate the same amount
of synthetic data that we use in our experiments. To
ensure a fair comparison, we use the same foreground masks and background images with added blending
artifacts for the generation of synthetic data.
We report detailed results over multiple IoU thresholds in Table~\ref{table:pascal_voc}, and some
qualitative results in Figure~\ref{fig:teaser-figure}.

As observed in~\cite{modeling-visual-context}, we note that adding data generated from~\cite{cut-paste-learn} 
to training leads to a drop in performance. We also
noticed that adding data generated by~\cite{modeling-visual-context} also leads to a drop in SSD performance. 
In contrast, our method improves SSD performance
by $2.7\%$ at AP$_{0.8}$.

\noindent \textbf{Quality of Synthetic Data:}
We develop another metric to evaluate the quality of synthetic data for the task of person
detection. A hardness metric is defined as $1-p$, where $p$ is the probability of the synthetic
composite image containing a person, according to the baseline SSD. We argue that if the baseline network is easily able to
detect the person in a composite image, then it is an easy example and may not boost the
network's performance when added to the training set. A similar metric has been proposed by previous
works ~\cite{unsupervised-hard-negative-mining,shrivastavaCVPR16ohem,Wang2017a,lrm} for evaluating the quality of real data.

In Figure~\ref{fig:hardness_of_data}, we compare the hardness of data generated by our approach to
to that of~\cite{modeling-visual-context,cut-paste-learn}.
The X-axis denotes the SSD
confidence and the Y-axis captures fraction of samples generated. We generate the same amount of
data with all methods and take an average over multiple experiment runs to produce this result.
As shown in Figure~\ref{fig:hardness_of_data}, we generate significantly harder examples than
~\cite{modeling-visual-context,cut-paste-learn}.
Please find more qualitative examples and experiments in the supplementary material.

\subsubsection{Ablation Studies}
\label{subsection:ablation_studies}

\begin{table}
\centering
\scalebox{0.78}{
  \begin{tabular}{c|r|r|r|r|r}
\hline
IoU  & Baseline & \cite{cut-paste-learn} & \cite{modeling-visual-context} & Ours no-$\mathcal{D}$ & Ours +$\mathcal{D}$ \\  \hline
$0.5$  &  $78.93$  &  $76.65$  &  $76.81$  &  $\mathbf{79.61}$ ($\mathbf{+0.68}$) &  $79.53$ ($+0.60$) \\ \hline
$0.6$  &  $69.61$  &  $66.88$  &  $66.91$  &  $70.39$ ($+0.78$) &  $\mathbf{70.67}$ ($\mathbf{+1.06}$) \\ \hline
$0.7$  &  $52.97$  &  $52.12$  &  $50.21$  &  $53.71$ ($+0.74$) &  $\mathbf{54.50}$ ($\mathbf{+1.53}$) \\ \hline
$0.8$  &  $29.54$  &  $28.82$  &  $28.14$  &  $31.96$ ($+2.44$) &  $\mathbf{32.22}$ ($\mathbf{+2.68}$) \\ \hline
\end{tabular}}
\caption{Results on VOC 2007 test data for person detection. 
Our augmentation improves the baseline over different IoU
thresholds by $~2.7\%$ at an IoU of $0.8$.}
\label{table:pascal_voc}
\end{table}
 
Table~\ref{table:ablation_studies} studies the effect of various parameters on the performance of an
SSD network fine-tuned on our data. In particular, we study the effect of (i) excluding noisy
foreground segmentation annotations during generation, (ii) using dropout in the synthesizer, (iii)
adding blending artifacts in the background, (iv) fine-tuning with real and synthetic data, and (v)
adding the discriminator. Our performance metric is mAP at an IoU threshold of $0.5$.
While we note progressive improvements in our performance with each addition, we see a slight drop
in performance after the addition of the discriminator. We investigate this further in
Table~\ref{table:pascal_voc}, and note that adding the discriminator improves our performance on
all IoU thresholds higher $>0.5$, allowing us to predict bounding boxes which are much better aligned with the ground truth boxes.

\begin{figure*}[t!]
\centering
{\includegraphics[width=0.8\textwidth,height=4.5in]{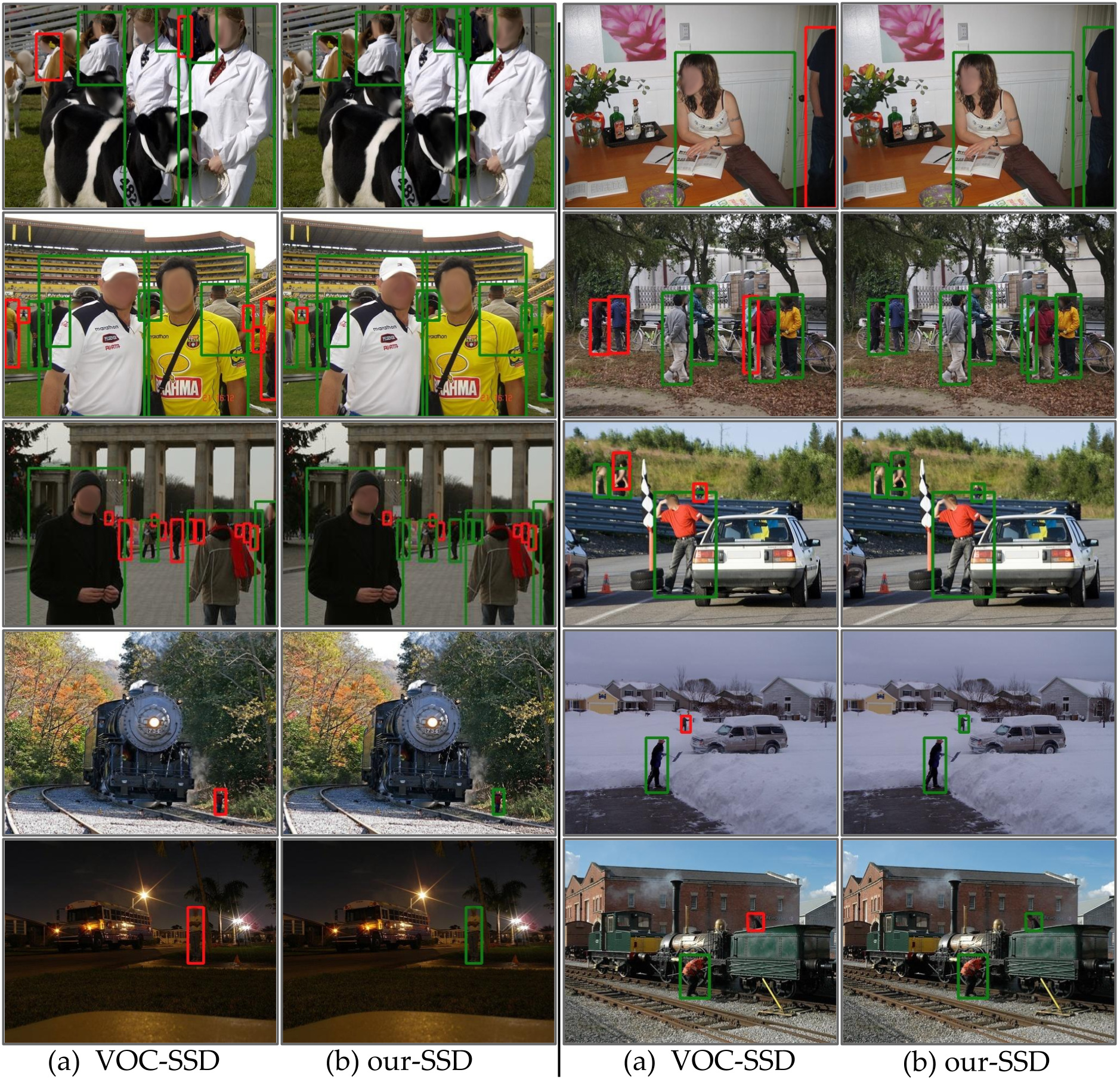}}
\caption{Qualitative results for VOC 2007 test set before and after training SSD with our
synthesized data. Green and red boxes show {\color{green}{correct}} and {\color{red}{missed}}
detections, respectively.
Note that synthetic data helps improve the SSD performance on severely occluded and small
instances.} 
\label{fig:ssd_qualitative_results_new}
\end{figure*}

\subsection{Experiments on GMU Data} 
\label{subsection:gmu_kitchen}
Lastly, we apply our data synthesis framework to improve the Faster-RCNN~\cite{FasterRCNN} for
instance detection. We compare our approach with baseline Faster-RCNN and the method
of~\cite{cut-paste-learn} on the GMU Kitchen Dataset~\cite{GMU-kitchen-dataset}.
%

The GMU Kitchen Dataset comprises $11$ classes and has 3-fold train/test splits as reported
in~\cite{cut-paste-learn}.
We use foregrounds from the Big Berkeley Instance Recognition (BigBIRD)~\cite{BigBird} dataset
and backgrounds from 
the UW Scenes dataset~\cite{UWScenes}.

Table~\ref{table:gmu_results} reports per class accuracy and mean average precision on the GMU test
set. Our approach out-performs baseline Faster-RCNN and~\cite{cut-paste-learn} by $3.5\%$ and $1\%$
in mAP, respectively. Interestingly, we improve accuracy of some categories such as
'palmolive-orange' by up to $7.5\%$.

\section{Conclusion}
\label{section:conclusion}
The recent success of deep learning has been fueled by supervised training requiring human
annotations. Large training sets are essential for improving performance under challenging real
world environments, but are difficult, expensive and time-consuming to obtain. Synthetic data
generation offers promising new avenues to augment training sets to improve the accuracy of deep neural networks.

In this paper, we introduced the concept of task-aware synthetic data generation to improve the
performance of a target network. Our approach trains a synthesizer to generate efficient and useful
synthetic samples, which helps to improve the performance of the target network. The target network provides feedback
to the synthesizer, to generate meaningful training samples. We proposed a novel approach to make
the target model invariant to blending artefacts by adding similar artefacts on background regions
of training images. We showed that our approach is efficient, requiring less number of samples as
compared to random data augmentations to reach a certain accuracy. In addition, we show a $2.7\%$
improvement in the state-of-art person detection using SSD. Thus, we believe that we have improved
the state-of-art in synthetic data generation tailored to improve deep learning techniques.

Our work opens up several avenues for future research. Our synthesizer network outputs affine
transformation parameters, but can be easily extended to output additional learnable photometric
transformations to the foreground masks and non-linear deformations. We showed composition
using a foreground and background image, but compositing multiple images can offer further
augmentations. While we showed augmentations in 2D using 2D cut-outs, our work can be extended to
paste rendered 3D models into 2D images. Our approach can also be extended to other target networks such as
regression and segmentation networks. Future work includes explicitly adding a diversity metric to
data synthesis to further improve its efficiency.

\section{Acknowledgements}
We wish to thank Kris Kitani for valuable discussions on this topic.

    



\newpage
{\small
\bibliographystyle{ieee_fullname}
\bibliography{egbib}
}

\begin{figure*}[t!]
	\centering
	\textbf{\large Supplementary Material: Learning to Generate Synthetic Data via Compositing}
\end{figure*}
\pagebreak

\appendix

\section{Implementation Details}
Our pipeline is implemented in python using the {PyTorch} deep-learning library. In the following subsections, we furnish relevant empirical details for our experiments on 
the different datasets in the manuscript.

\subsection{Experiments on AffNIST data}

In our experiments on the AffNIST benchmark, our synthesizer network generates affine transformations which are applied to MNIST images. These transformed images are then used to augment the MNIST training set and the performance of a handwritten digit classifier network trained on the augmented dataset is compared to an equivalent classifier trained on MNIST + AffNIST images.
In the rest of the section, we give details about our experimental settings.\\



\noindent \textbf{Data Pre-processing.}
For the AffNIST benchmark, our synthesizer network uses foreground masks from the MNIST training dataset. 
We pad the original $28 \times 28$ MNIST images by $6$ black pixels on all sides to enlarge them to a $40 \times 40$ resolution.
This is done to ensure that images generated by our synthesizer have the same size as the AffNIST images which also have a spatial resolution of $40 \times 40$ pixels. \\

\noindent \textbf{Architecture of the Synthesizer Network.}
The synthesizer network takes as input a foreground image from the MNIST dataset and outputs 
a $6-$dimensional vector representing the affine parameters, namely the (i) angle of rotation, (ii) translation along the X$-$axis,
(iii) translation along the Y$-$axis, (iv) shear, (v) scale along the X$-$axis, and (vi) scale along the Y$-$axis.
We clamp these parameters to the same range used to generate AffNIST dataset. 
These parameters are used to define an affine transformation matrix which is fed to a Spatial Transformer module \cite{spatial-transformer-networks} alongside
the foreground mask. The Spatial Transformer module applies the transformation to the foreground mask and returns the synthesized image. We attach the pytorch model dump for the
synthesizer below.

\begin{lstlisting}
Features(
  (BackBone): Sequential(
    (0): C2d(1, 10, ksz=(5, 5), st=(1, 1))
    (1): MaxPool2d(ksz=(3, 3), st=(2, 2))
    (2): BN2d(10, eps=1e-05, mmntm=0.1)
    (3): ReLU(inplace)
    (4): Dropout2d(p=0.5)
  )
  (FgBranch): Sequential(
    (0): C2d(10, 20, ksz=(3, 3), st=(1, 1))
    (1): ReLU(inplace)
    (2): BN2d(20, eps=1e-05, mmntm=0.1)
    (3): Dropout2d(p=0.5)
    (4): C2d(20, 20, ksz=(3, 3), st=(1, 1))
    (5): ReLU(inplace)
    (6): BN2d(20, eps=1e-05, mmntm=0.1)
    (7): Dropout2d(p=0.5)
  )
)
RegressionFC(
  (features): Sequential(
    (0): C2d(40, 20, ksz=(3, 3), st=(1, 1))
    (1): ReLU(inplace)
    (2): BN2d(20, eps=1e-05, mmntm=0.1)
    (3): Dropout2d(p=0.5)
    (4): C2d(20, 20, ksz=(3, 3), st=(1, 1))
    (5): ReLU(inplace)
    (6): BN2d(20, eps=1e-05, mmntm=0.1)
    (7): Dropout2d(p=0.5)
  )
  (regressor): Sequential(
    (0): Linear(in_f=1620, out_f=50, bias=True)
    (1): ReLU(inplace)
    (2): BatchNorm1d(50, eps=1e-05, mmntm=0.1)
    (3): Dropout(p=0.5)
    (4): Linear(in_f=50, out_f=20, bias=True)
    (5): ReLU(inplace)
    (6): BatchNorm1d(20, eps=1e-05, mmntm=0.1)
    (7): Dropout(p=0.5)
    (8): Linear(in_f=20, out_f=6, bias=True)
  )
)
\end{lstlisting}

The acronyms used in the pytorch model dump are described in Table.~\ref{tab:acronym}.
\begin{table}[h]
  \centering
  \begin{tabular}{r|l}
    \hline
    \textbf{Acronym} & \textbf{Meaning} \\
    \hline
    C2d & Conv2d \\ \hline
    BN2d & BatchNorm2d \\ \hline
    ksz & kernel\_size \\ \hline
    st & stride \\ \hline
    pdng & padding \\ \hline
    LReLU & LeakyReLU \\ \hline
    neg\_slp & negative\_slope \\ \hline
    InstNrm2D & InstanceNorm2d \\ \hline
    mmntm & momentum\\ \hline
    in\_f & in\_features \\ \hline
    out\_f & out\_features \\ \hline
\end{tabular}
\caption{Acronyms used within PyTorch model dumps.}
\label{tab:acronym}
\end{table}

\noindent \textbf{Architecture of the Target Classifier.}
For our experiments in Figure.~7, we use a target model with two convolutional layers followed by a dropout layer, and two linear layers.
The pytorch model dump is attached below.\\

\begin{lstlisting}
MNISTClassifier(
  (0): C2d(1, 10, ksz=(5, 5), st=(1, 1))
  (1): ReLU(inplace)
  (2): C2d(10, 20, ksz=(5, 5), st=(1, 1))
  (3): ReLU(inplace)
  (4): Dropout2d(p=0.5)
  (5): Linear(in_f=980, out_f=50, bias=True)
  (6): ReLU(inplace)
  (7): Linear(in_f=50, out_f=10, bias=True)
)
\end{lstlisting}

For our experiments in Table.~1, we use the following architecture from \cite{zhao2018towards}, where Swish denotes the swish activation function from \cite{swish}.
\begin{lstlisting}
MNISTClassifierDeep(
  (0): C2d(1, 64, ksz=(5, 5), st=(2, 2))
  (1): BN2d(64, eps=1e-05, mmntm=0.1)
  (2): Swish(inplace)
  (3): C2d(64, 64, ksz=(5, 5), st=(2, 2))
  (4): BN2d(64, eps=1e-05, mmntm=0.1)
  (5): Swish(inplace)
  (6): C2d(64, 64, ksz=(5, 5), st=(2, 2))
  (7): BN2d(64, eps=1e-05, mmntm=0.1)
  (8): Swish(inplace)
  (9): C2d(64, 64, ksz=(5, 5), st=(2, 2))
  (10): BN2d(64, eps=1e-05, mmntm=0.1)
  (11): Swish(inplace)
  (12): Linear(in_f=576, out_f=50, bias=True)
  (13): Swish(inplace)
  (14): Linear(in_f=50, out_f=10, bias=True)
)
\end{lstlisting}
\noindent \textbf{Training the Synthesizer Network}. We use the ADAM optimizer with a batch-size of $1024$ and a fixed
learning rate of $10^{-3}$. Xavier initialization is used with a gain of 0.4 to initialize
the network weights. The synthesizer is trained in lock-step with the target model: we alternately update the synthesizer and target models.
The weights of the target model are fixed during the synthesizer training. The synthesizer is trained until we find $500$ hard examples
per class. A synthesized image is said to be a hard example if $p - p^* > 0.05$ where $p*$ is the probability of it belonging to the ground truth class,
and $p$ is the maximum probability over the other classes estimated by the target model. 
We maintain a cache of hard examples seen during the synthesizer training and use this cache for training dataset augmentation. 
We observed from our experiments that the number of epochs required to generated 500 images per class increases over sleep cycles as the target model
became stronger. \\

\noindent \textbf{Training the Target Network}. The original training dataset consists of MNIST training images.
After each phase of synthesizer network training we augment the training dataset with all images from the cache of hard examples and train the target network for $30$ epochs.
The augmented dataset is used to fine-tune the classifier network. For fine-tuning, we use SGD optimizer with a batch-size of 64, 
$10^{-2}$ learning rate and a momentum of $0.5$. For our experiments in Table. $1$, we reduce the learning rate to $10^{-3}$ after $100$ iterations of training the synthesizer and target networks. 

\subsection{Experiments on Pascal VOC data}
\label{app:pascal}
In our experiments on the Pascal VOC person detection benchmark, we additionally employ a natural versus synthetic image
discriminator to encourage the synthetic images to appear realistic. We use the SSD$-300$ pipeline from \cite{SSD-network} as the target model.\\

\begin{figure}
	\centering
	\includegraphics[width=80mm,height=20mm]{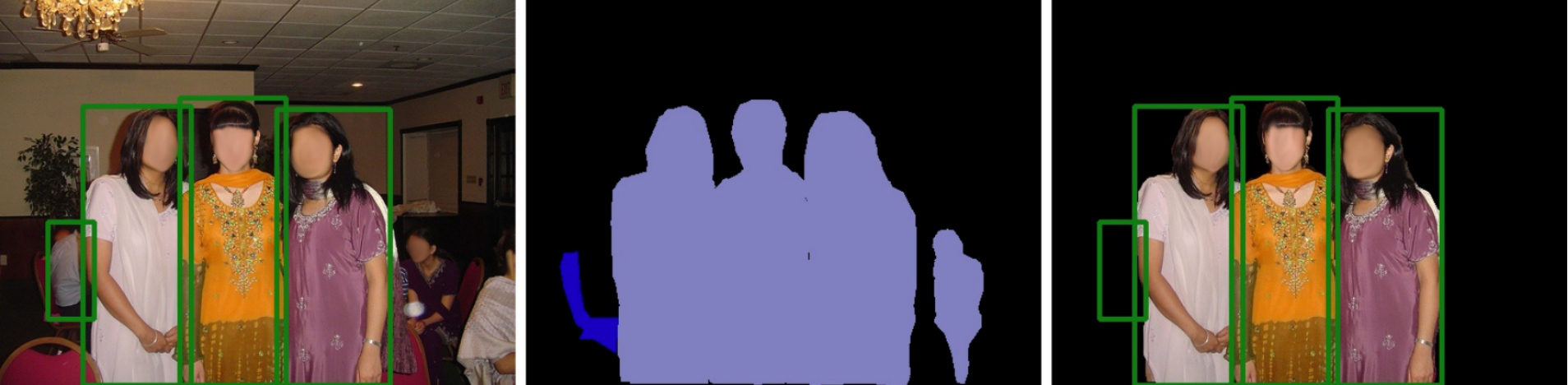}\\
	\includegraphics[width=80mm,height=20mm]{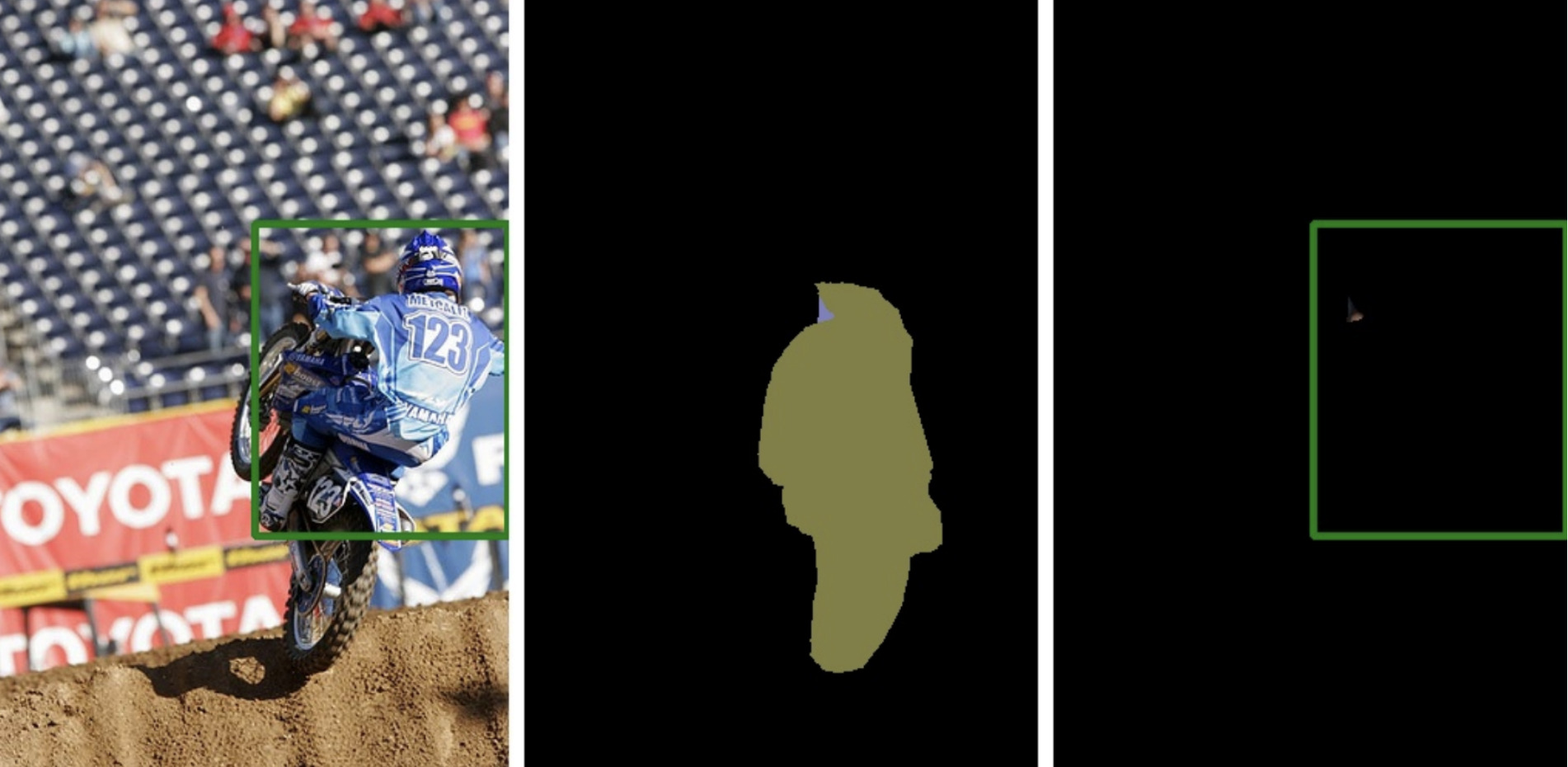}\\
	\includegraphics[width=80mm,height=30mm]{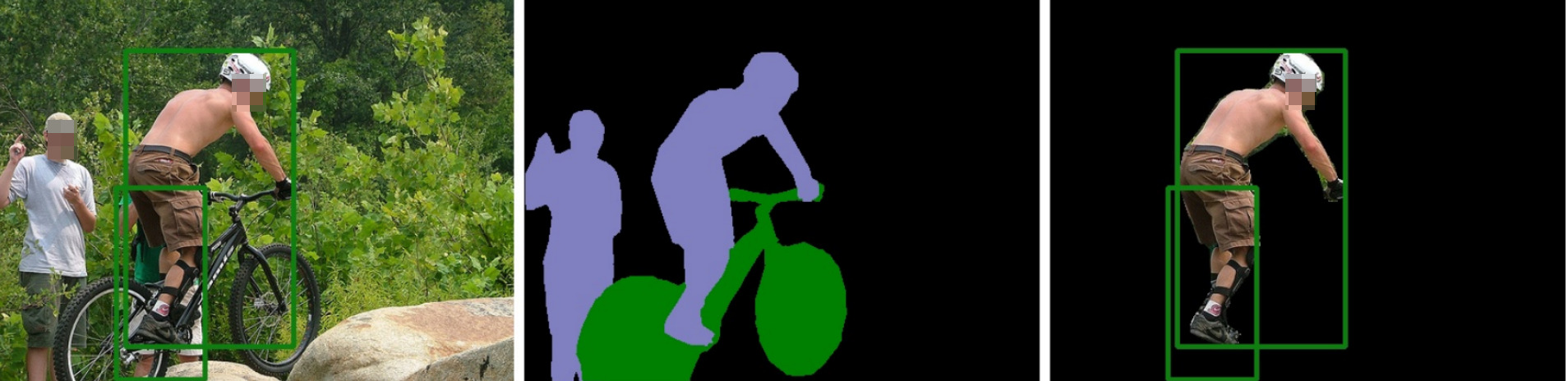}
	\caption{Disagreements between segmentation and detection annotations in the VOC dataset.
		Column $1$ shows ground truth bounding boxes overlaid on images. Column $2$ shows
		segmentation annotations. Column $3$ shows generated instance masks: each green box corresponds to one instance.}
	\label{fig:annotation_errors}
\end{figure}

\noindent \textbf{Data Pre-processing}. We resize all synthetic images to $300 \times 300$ pixels to be consistent with the SSD$-300$ training protocol. 
Background images for generating synthetic data are drawn from the COCO dataset, and foreground masks come
from the VOC 2007 and 2012 trainval datasets. VOC datasets contain instance segmentation masks for a subset of trainval images.
We use the ground truth segmentation masks and bounding box annotations to recover additional instance segmentation masks.
We noticed that for about $10\%$ images the segmentation and bounding box detections do not agree, therefore we visually inspected the 
instance masks generated and filtered out erroneous ones. Figure~\ref{fig:annotation_errors} shows some images from the VOC dataset where
the segmentation and bounding box annotations disagree, resulting in erroneous instance masks.

The foreground segmentations are centered
and normalized such that $\max($height, width$)$ of the segmentation bounding-box occupies at least
$0.7$ of the corresponding image dimension. We randomly pair background images with foreground instances while training
the synthesizer network and during synthetic data generation.\\

\noindent \textbf{Architecture of the Synthesizer Network}. The synthesizer architecture is similar to the one
used for AffNIST experiments with two enhancements: (i) we use the fully convolutional part of the the VGG$-16$~\cite{vgg} network as the backbone, 
and (ii) we have an additional mid-level feature extraction subnetwork for the background image.
This is described in the following model dump.

\begin{lstlisting}
Features( 
  (BackBone): Sequential( 
    (0): C2d(3, 64, ksz=(3, 3), pdng=(1, 1)) 
    (1): BN2d(64, eps=1e-05, mmntm=0.1) 
    (2): ReLU(inplace) 
    (3): C2d(64, 64, ksz=(3, 3), pdng=(1, 1)) 
    (4): BN2d(64, eps=1e-05, mmntm=0.1) 
    (5): ReLU(inplace)
    (6): MaxPool2d(ksz=2, st=2) 
    (7): C2d(64, 128, ksz=(3, 3), pdng=(1, 1)) 
    (8): BN2d(128, eps=1e-05, mmntm=0.1) 
    (9): ReLU(inplace) 
    (10): C2d(128, 128, ksz=(3, 3), pdng=(1, 1)) 
    (11): BN2d(128, eps=1e-05, mmntm=0.1) 
    (12): ReLU(inplace) 
    (13): MaxPool2d(ksz=2, st=2) 
    (14): C2d(128, 256, ksz=(3, 3), pdng=(1, 1)) 
    (15): BN2d(256, eps=1e-05, mmntm=0.1) 
    (16): ReLU(inplace) 
    (17): C2d(256, 256, ksz=(3, 3), pdng=(1, 1)) 
    (18): BN2d(256, eps=1e-05, mmntm=0.1) 
    (19): ReLU(inplace) 
    (20): C2d(256, 256, ksz=(3, 3), pdng=(1, 1)) 
    (21): BN2d(256, eps=1e-05, mmntm=0.1) 
    (22): ReLU(inplace) 
    (23): MaxPool2d(ksz=2, st=2) 
    (24): C2d(256, 512, ksz=(3, 3), pdng=(1, 1)) 
    (25): BN2d(512, eps=1e-05, mmntm=0.1)
    (26): ReLU(inplace) 
    (27): C2d(512, 512, ksz=(3, 3), pdng=(1, 1)) 
    (28): BN2d(512, eps=1e-05, mmntm=0.1) 
    (29): ReLU(inplace) 
    (30): C2d(512, 512, ksz=(3, 3), pdng=(1, 1)) 
    (31): BN2d(512, eps=1e-05, mmntm=0.1) 
    (32): ReLU(inplace) 
    (33): MaxPool2d(ksz=2, st=2) 
  ) 
  (FgBranch): Sequential( 
    (0): C2d(512, 256, ksz=(3, 3), st=(1, 1)) 
    (1): BN2d(256, eps=1e-05, mmntm=0.1) 
    (2): ReLU(inplace) 
    (3): C2d(256, 256, ksz=(3, 3), st=(1, 1)) 
    (4): BN2d(256, eps=1e-05, mmntm=0.1) 
    (5): ReLU(inplace) 
    (6): C2d(256, 20, ksz=(3, 3), st=(1, 1)) 
  ) 
  (BgBranch): Sequential( 
    (0): C2d(512, 256, ksz=(3, 3), st=(1, 1)) 
    (1): BN2d(256, eps=1e-05, mmntm=0.1) 
    (2): ReLU(inplace) 
    (3): C2d(256, 256, ksz=(3, 3), st=(1, 1)) 
    (4): BN2d(256, eps=1e-05, mmntm=0.1) 
    (5): ReLU(inplace) 
    (6): C2d(256, 20, ksz=(3, 3), st=(1, 1)) 
  ) 
) 
RegressionFC(
  (features): Sequential( 
    (0): C2d(40, 64, ksz=(5, 5), pdng=(2, 2)) 
    (1): ReLU(inplace) 
    (2): BN2d(64, eps=1e-05, mmntm=0.1) 
    (3): C2d(64, 64, ksz=(5, 5), pdng=(2, 2)) 
    (4): ReLU(inplace) 
    (5): BN2d(64, eps=1e-05, mmntm=0.1) 
  ) 
  (regressor): Sequential( 
    (0): Linear(in_f=64, out_f=128, bias=True)
    (1): ReLU(inplace) 
    (2): BatchNorm1d(128, eps=1e-05, mmntm=0.1) 
    (3): Linear(in_f=128, out_f=128, bias=True) 
    (4): ReLU(inplace) 
    (5): BatchNorm1d(128, eps=1e-05, mmntm=0.1) 
    (6): Linear(in_f=128, out_f=6, bias=True) 
  ) 
)
\end{lstlisting}

\noindent \textbf{Architecture of the Discriminator}. 
Our discriminator is based on the discriminator architecture from \cite{pix2pixHD}. 
\begin{lstlisting}
Discriminator(
  (0): Sequential(
    (0): C2d(3, 64, ksz=(4, 4), st=(2, 2), pdng=(2, 2))
    (1): LReLU(neg_slp=0.2, inplace) 
  ) 
  (1): Sequential( 
    (0): C2d(64, 128, ksz=(4, 4), st=(2, 2), pdng=(2, 2)) 
    (1): InstNrm2D(128, eps=1e-05, mmntm=0.1, affine=False, track_running_stats=False) 
    (2): LReLU(neg_slp=0.2, inplace) 
  ) 
  (2): Sequential( 
    (0): C2d(128, 256, ksz=(4, 4), st=(2, 2), pdng=(2, 2)) 
    (1): InstNrm2D(256, eps=1e-05, mmntm=0.1, affine=False, track_running_stats=False) 
    (2): LReLU(neg_slp=0.2, inplace) 
  ) 
  (3): Sequential( 
    (0): C2d(256, 512, ksz=(4, 4), st=(1, 1), pdng=(2, 2)) 
    (1): InstNrm2D(512, eps=1e-05, mmntm=0.1, affine=False, track_running_stats=False) 
    (2): LReLU(neg_slp=0.2, inplace) 
  ) 
  (4): Sequential( 
    (0): C2d(512, 1, ksz=(4, 4), st=(1, 1), pdng=(2, 2)) 
  ) 
  (5): AvgPool2d(ksz=3, st=2, pdng=[1, 1]) 
) 
\end{lstlisting}
 
\noindent \textbf{Training the Synthesizer, Discriminator and Target Networks}. 
The synthesizer is trained in lock-step with the discriminator and the target models: the discriminator
and target model weights are fixed and the synthesizer is trained for $1000$ batches. The synthesizer is
then used to generate synthetic images: we do a forward pass on $100$ batches of randomly paired foreground and background images and
pick composite images which have a target estimated probability of less than $0.5$. These hard examples are added to 
the training dataset. The weights of the synthesizer network are then fixed and the discriminator and target models are
trained for $3$ epochs over the training set. This cycle is repeated $5$ times.

The VGG-16 backbone of the synthesizer is initialized with an ImageNet pretrained model.
The synthesizer is trained using the ADAM optimizer with a batch-size of $16$ and a fixed
learning rate of $1e-4$ is used. The weights of the discriminator are randomly initialized to a Normal distribution with a variance of $0.02$.
The discriminator is trained using the ADAM optimizer with 
a batch size of $16$ and a fixed learning rate of $1e-4$ is used. We initialize the SSD model with the final weights from \cite{SSD-network}.
The SSD model is trained using the SGD optimizer, a fixed learning rate of $1e-5$, momentum of $0.9$ and a weight decay of $0.0005$. \\

\noindent \textbf{Custom Spatial Transformer}. The image resizing during data pre-processing step and the
spatial-transformer module both involve bilinear interpolation which introduces quantization artefacts
near the segmentations mask edges. To deal with these artefacts we customize the Spatial Transformer implementation.
More specifically, we remove these artefacts by subtracting a scalar($\mu=1-10^{-7}$)
from the segmentation masks and applying ReLU non-linearity. The result is normalized back to a
binary image. Gaussian blur is applied to the resulting mask for smooth transition near edges before
applying alpha blending.

\subsection{Experiments on GMU Kitchen data}
Our experiments on the GMU data use the same synthesizer architecture and training strategy as described in Section.~\ref{app:pascal} with one change: we do not
use the discriminator in these experiments. We did not see noticeable improvements in performance with the addition of the discriminator in these experiments.

\section{Qualitative Results}
Qualitative results follow on the next page.

\begin{figure*}[t!]
\centering
\includegraphics[width=\textwidth,height=6in]{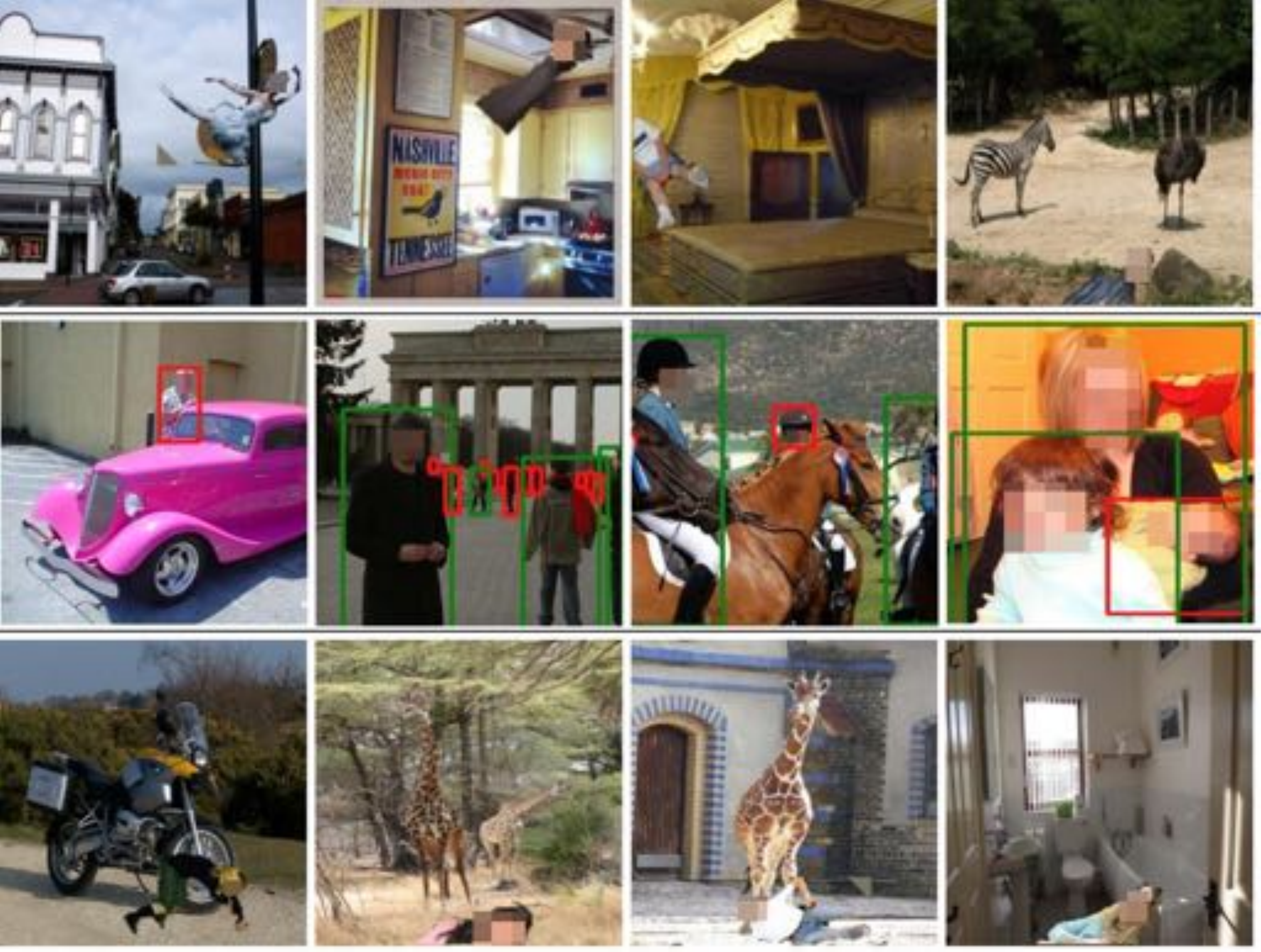}
\caption{
Qualitative results for Synthetic Data Generation.
\textbf{Row-1}: Here we show the synthetic images generated by our synthesizer based on feedback from the baseline SSD trained on the VOC 2007-2012 dataset. These images
evoke misclassifications from the baseline SSD primarily because they present human instances in unforeseen/unrealistic circumstances. \textbf{Row-2}: SSD failures on VOC 2007 test images after finetuning the baseline SSD with composite images such as those shown in Row-1. We notice the the three failure cases are a) person instances at small scales, b) horizontal pose, and c) severe occlusion. \textbf{Row-3}: Synthetic images generated by our synthesizer after feedback from the finetuned SSD. We notice that our synthesizer now generates small scales, horizontal pose and severely truncated instances.}
\end{figure*}

\clearpage
\begin{figure*}[t!]
\centering
\includegraphics[width=0.95\textwidth]{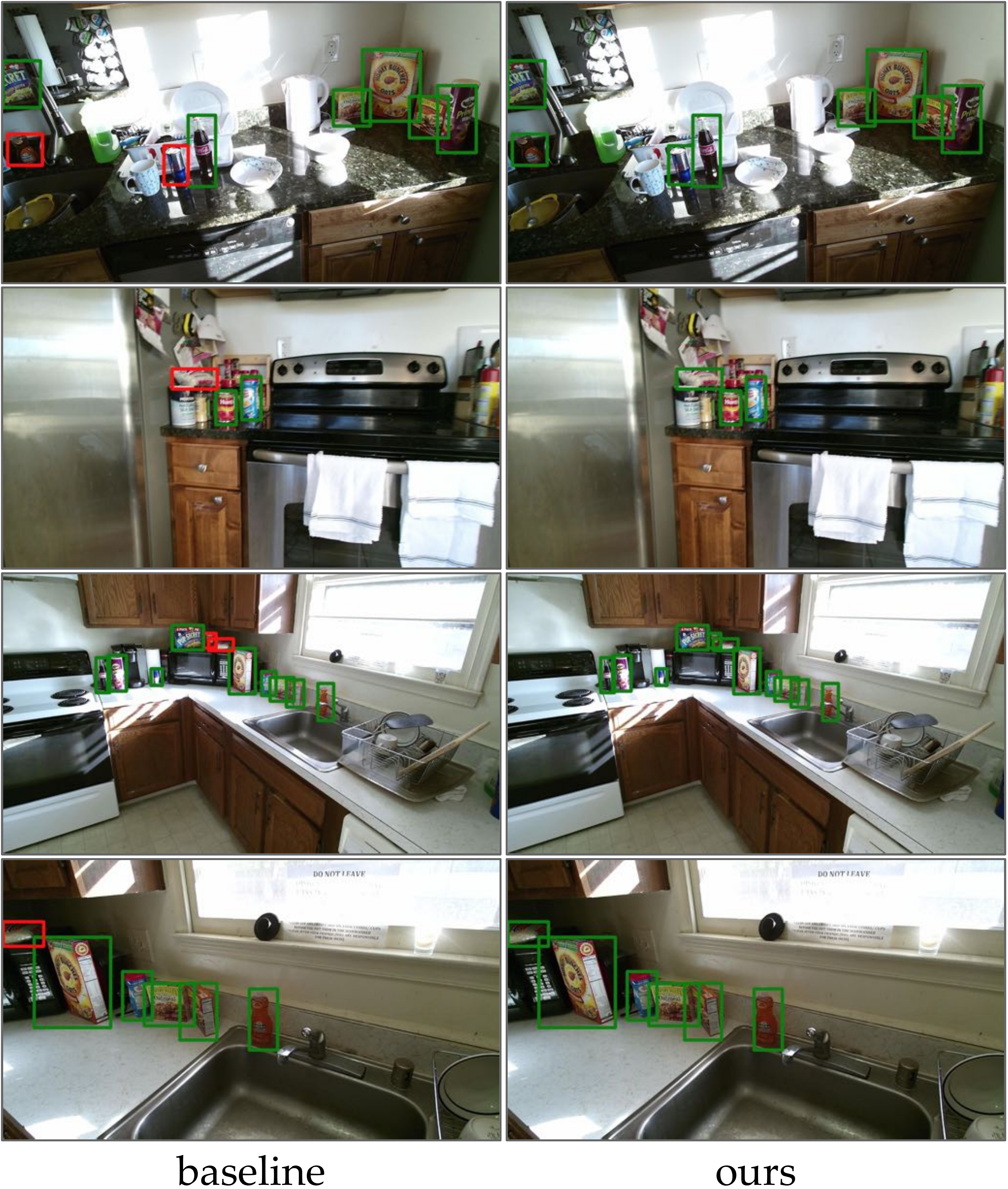}
\caption{Qualitative improvements on the GMU Kitchen benchmark. Green and red bounding boxes denote correct and missing detections respectively.}
\end{figure*}

\clearpage
\begin{figure*}[t!]
\centering
\includegraphics[width=0.95\textwidth]{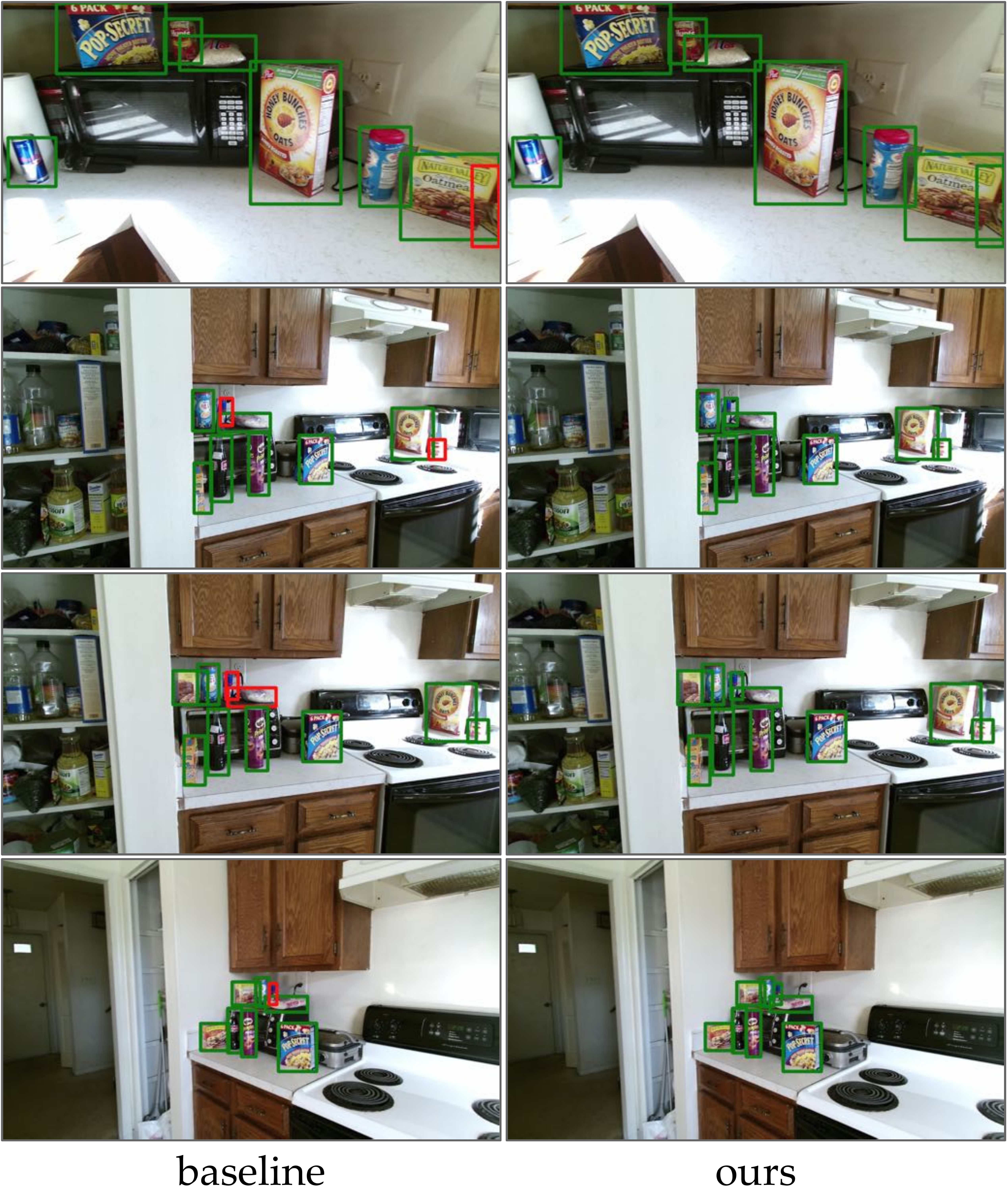}
\caption{More qualitative improvements on the GMU Kitchen benchmark. Green and red bounding boxes denote correct and missing detections respectively.}
\end{figure*}

\clearpage
\begin{figure*}[t!]
\centering
\includegraphics[width=0.95\textwidth]{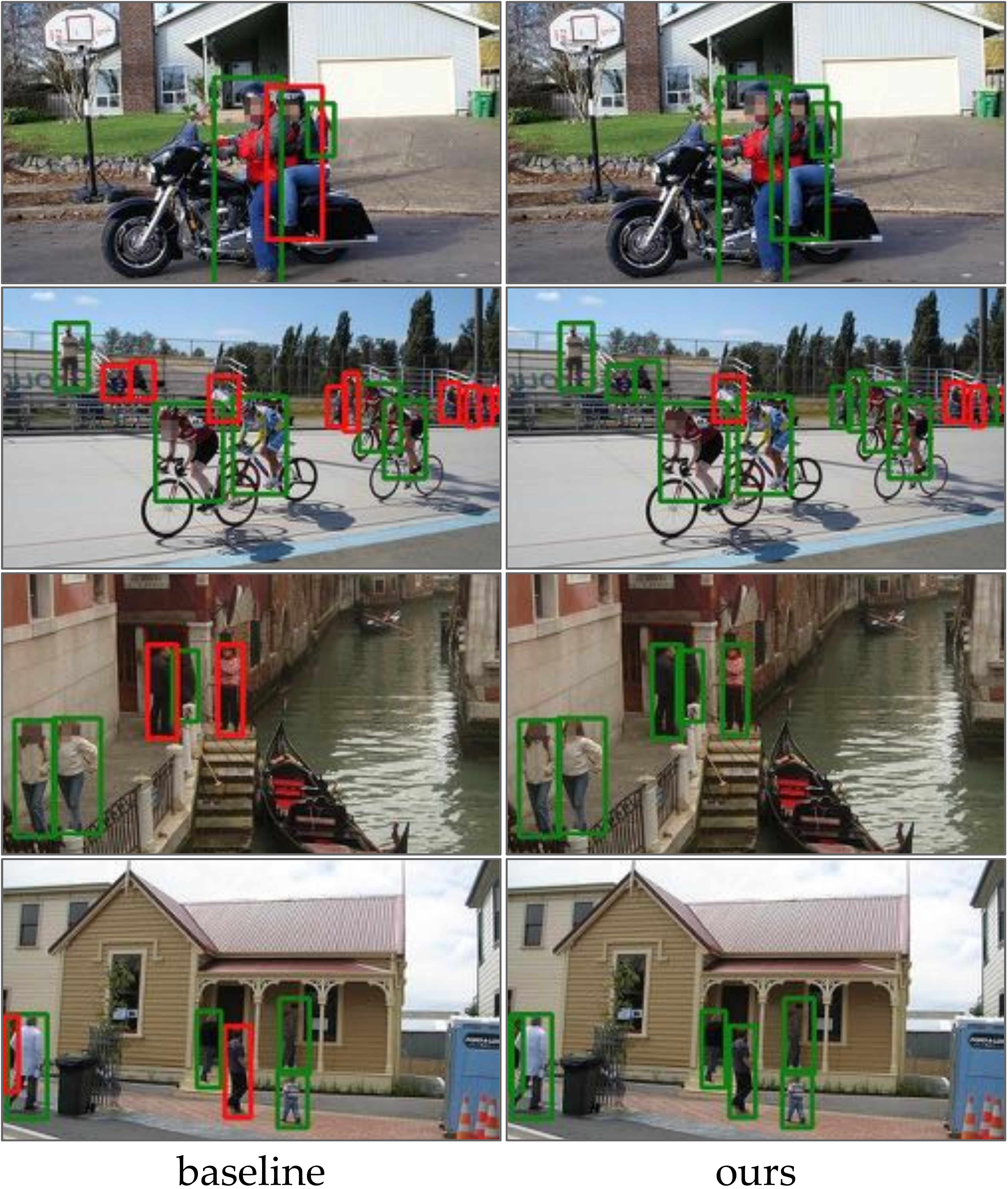}
\caption{Qualitative improvements on the Pascal VOC person detection benchmark. Green and red bounding boxes denote correct and missing detections respectively.}
\end{figure*}

\clearpage
\begin{figure*}[t!]
\centering
\includegraphics[width=0.95\textwidth]{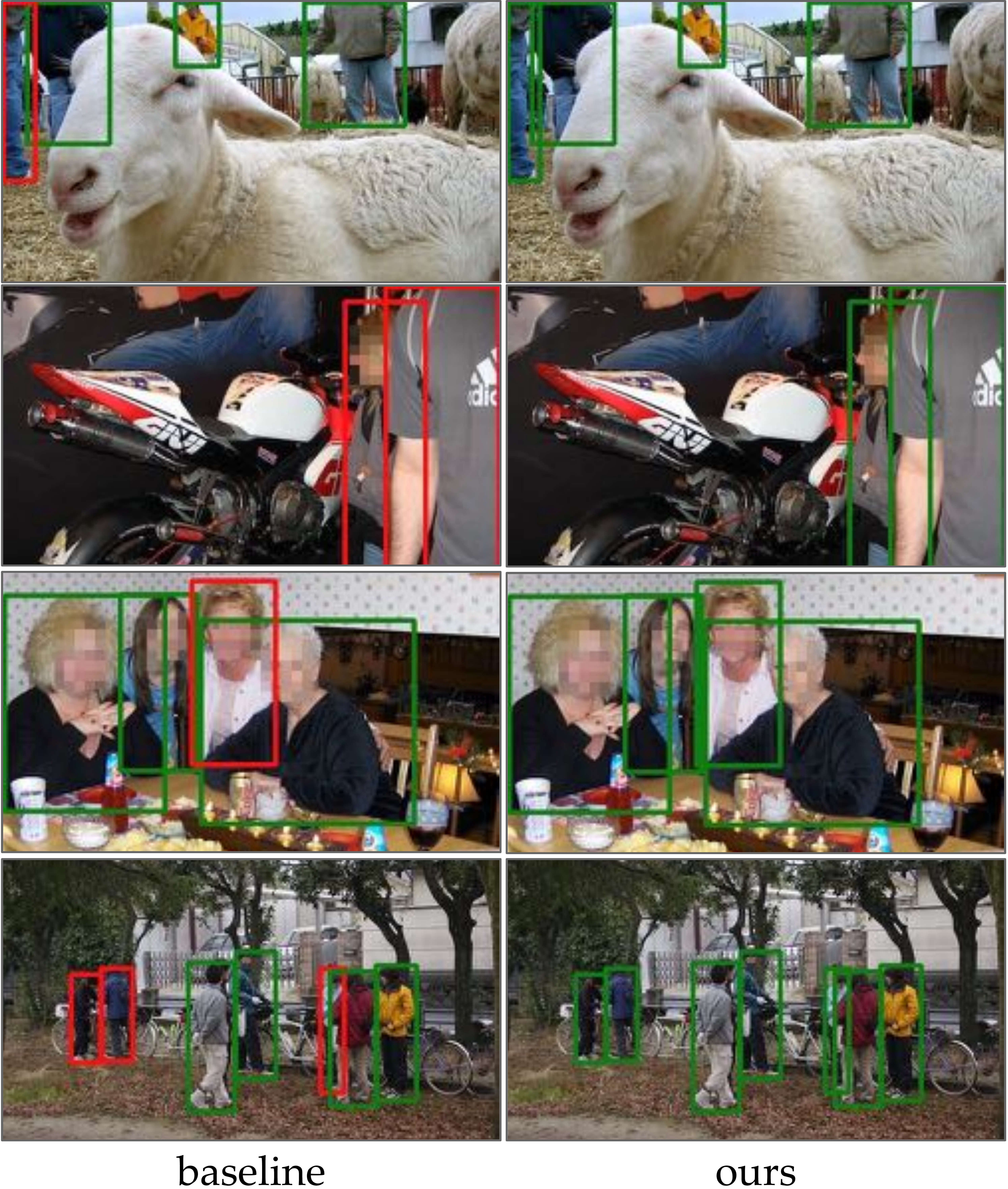}
\caption{More qualitative improvements on the Pascal VOC person detection benchmark. Green and red bounding boxes denote correct and missing detections respectively.}
\end{figure*}

\clearpage

\end{document}